\documentclass[12pt,logo]{spell}
\usepackage{nicefrac}           

\usepackage{multirow}
\usepackage{array}
\usepackage{adjustbox}
\usepackage{makecell}
\usepackage{longtable}
\usepackage{arydshln}           

\usepackage{wrapfig}
\usepackage{subcaption}
\usepackage{rotating}
\usepackage[abs]{overpic}
\usepackage{tikz}               

\usepackage[most]{tcolorbox}

\usepackage{listings}
\usepackage{csquotes}

\usepackage{fontawesome}

\usepackage[capitalize]{cleveref}

\usepackage{tocloft}

\usepackage{lipsum}

\definecolor{bluelink}{RGB}{0,113,188}
\definecolor{greenlink}{RGB}{0,188,113}
\definecolor{linkteal}{RGB}{0,102,102}
\definecolor{linkblue}{RGB}{0,51,102}
\hypersetup{
    colorlinks=true,
    citecolor=green!93!black,
    linkcolor=linkblue,
    urlcolor=bluelink
}

\definecolor{codekeyword}{rgb}{0.0, 0.0, 0.5}
\definecolor{codecomment}{rgb}{0.0, 0.5, 0.0}
\definecolor{codestring}{rgb}{0.56, 0.0, 1.0}

\lstdefinestyle{pythonstyle}{
    language=Python,
    basicstyle=\ttfamily\small,
    keywordstyle=\color{codekeyword}\bfseries,
    commentstyle=\color{codecomment}\itshape,
    stringstyle=\color{codestring},
    showstringspaces=false,
    breaklines=true,
    tabsize=4,
    numbers=none,
    frame=none,
    backgroundcolor=\color{white},
    captionpos=b,
    morekeywords={self, __init__, __name__, __main__},
}
\title{\center Do VLMs Need Vision Transformers?\\Evaluating State Space Models as Vision Encoders}

\author{Shang-Jui Ray Kuo, Paola Cascante-Bonilla\\
    \vspace{-12pt}
    {\normalfont\small\texttt{\{skuo, paola\}@cs.stonybrook.edu}}\\
    \vspace{8pt}
    Stony Brook University\\
    \vspace{8pt}
    {\small
    \faGlobe\ \href{https://lab-spell.github.io/vlm-ssm-vision-encoders/}{Project Page} \quad
    \faGithub\ \href{https://github.com/raykuo18/vlm-ssm-vision-encoders}{Repository}
    }
}

\begin{abstract}
Large vision--language models (VLMs) often use a frozen vision backbone, whose image features are mapped into a large language model through a lightweight connector.
While transformer-based encoders are the standard visual backbone,
we ask whether state space model (SSM) vision backbones can be a strong alternative.
We systematically evaluate SSM vision backbones for VLMs in a controlled setting. Under matched ImageNet-1K initialization, the SSM backbone achieves the strongest overall performance across both VQA and grounding/localization. We further adapt both SSM and ViT-family backbones with detection or segmentation training and find that dense-task tuning generally improves performance across families; after this adaptation, the SSM backbone remains competitive while operating at a substantially smaller model scale.
We further observe that (i) higher ImageNet accuracy or larger backbones do not reliably translate into better VLM performance, and (ii) some visual backbones are unstable in localization. Based on these findings, we propose stabilization strategies that improve robustness for both backbone families and highlight SSM backbones as a strong alternative to transformer-based vision encoders in VLMs.
\end{abstract}

\begin{document}

\maketitle

\begin{figure}[t!]
  \centering
  \includegraphics[width=1\linewidth]{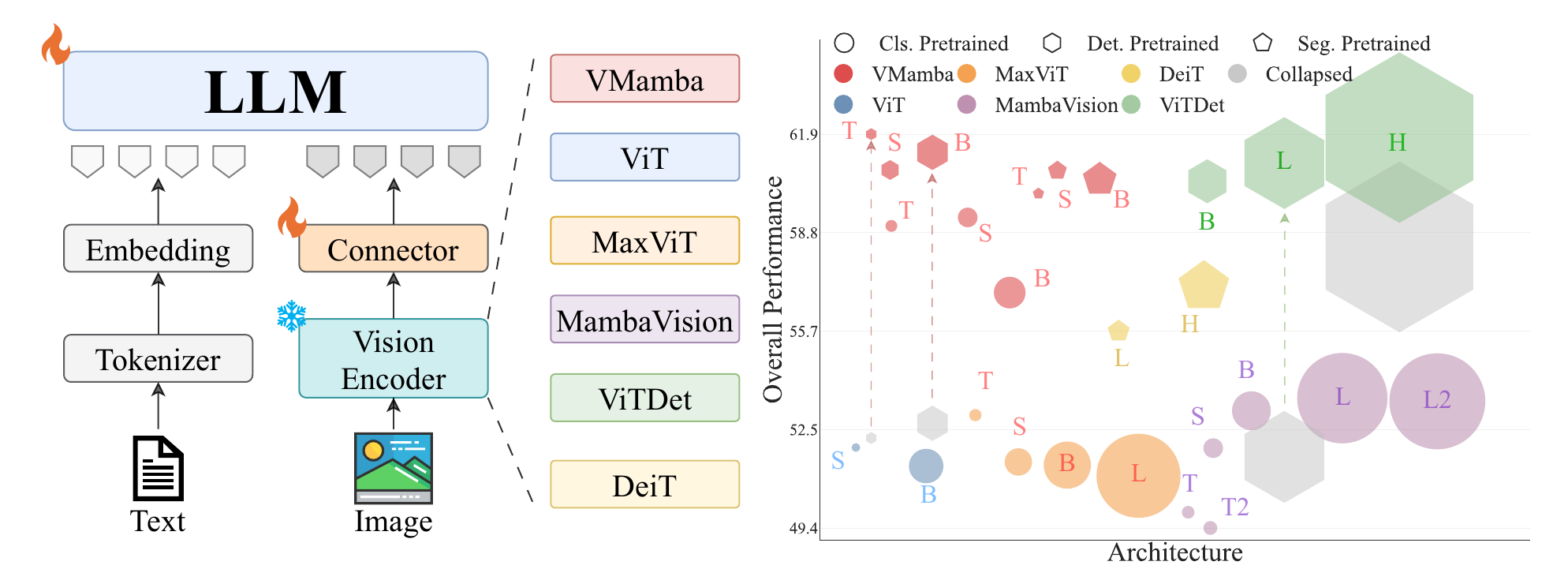}
  \vspace{-0.5cm}
  \caption{\textbf{Overview of our controlled vision encoder study.}
    We follow a LLaVA-style VLM design where an input image is encoded by a frozen vision encoder into visual tokens, which are then processed by the LLM; this modular setup enables controlled swaps of vision encoders from different architecture families under a fixed training recipe.
    \textbf{Right:} Each marker summarizes one evaluated vision-backbone checkpoint plugged into the same VLM setting. Colors denote the backbone family; marker shapes denote the pretraining objective (classification, detection, or segmentation); marker size reflects encoder scale. Gray markers indicate configurations that exhibit collapse; arrows point to the corresponding stabilized variants after applying our stabilizations.
    }
  \label{fig:figure1}
\end{figure}

\section{Introduction}
\label{sec:intro}

Recent progress in vision--language models (VLMs) commonly follows a modular design: a pretrained vision encoder produces visual tokens, a lightweight connector maps them into the embedding space of a large language model (LLM), and the combined system is instruction-tuned for open-ended generation~\cite{llava,blip2,mllm_survey}.
Typically, the vision encoder is kept fixed during instruction tuning, updating only the connector and the LLM~\cite{prismatic,cobra,qwen3vl}.
While some systems do finetune the vision encoder, doing so reliably requires careful optimization choices and can be unstable under standard instruction-tuning recipes~\cite{internvl35}; moreover, it obscures controlled comparisons of vision backbones by entangling architectural effects with training dynamics.
Therefore, freezing the vision backbone enables different backbones to be evaluated under matched multimodal training without entangling architectural effects with joint vision-language optimization~\cite{llava_more,cambrian,prismatic}.

Despite extensive work on VLM training recipes, the vision encoder remains relatively narrow in architectural choice.
Most systems still rely on ViT-family~\cite{vit}, or broadly transformer-based encoders, as the vision backbone~\cite{prismatic}.
At the same time, many comparisons change multiple ingredients together, including the vision pretraining objective, the multimodal training pipeline, resolution and tokenization settings, and connector design.
This makes it difficult to isolate what is due to the vision architecture itself and whether the choice of backbone family limits how much useful evidence can be delivered from the vision encoder~\cite{cambrian,llava_more}.

Another recurring challenge in VLMs is extracting spatially grounded evidence from images under a fixed multimodal token budget~\cite{nuwa,spatialrgpt}.
To capture fine details, VLMs often increase image resolution or the number of visual tokens, but this quickly raises compute and memory costs in both the vision encoder and the LLM~\cite{inferenceoptimalvlms,internvl35}. 
This limitation raises an interesting open question: \textit{{is there a better visual representation that also encodes richer spatial information without increasing the number of vision tokens?}}

State-space model (SSM) vision backbones have recently shown promising results in vision tasks, including competitive classification and particularly strong performance on dense prediction tasks~\cite{vmamba,vim,mambavision}.
Unlike ViTs, which rely on global self-attention over a flattened token sequence, many SSM vision models build representations through structured state-space updates implemented as multi-directional scans over the 2D grid~\cite{vmamba,2dmamba,spatialmamba}.
These properties make SSM backbones plausible candidates for producing visual tokens that retain fine spatial information, which is important when the LLM must reason over localized details.
To our knowledge, prior VLM work has not performed controlled backbone swaps that include SSM vision encoders while matching both the training recipe and the vision–language interface.

In this work, 
we use VMamba~\cite{vmamba} as a strong pure-SSM backbone baseline, given its 2D-Selective-Scan (SS2D) design, which shows strong performance on dense vision tasks.
Starting from ImageNet-1K (IN1K)~\cite{imagenet} supervised initialization, we compare vision backbones in a controlled LLaVA-style setting~\cite{llava}, and then adapt SSM and ViT backbones with detection or segmentation pretraining.
Building on this \textit{backbone}-controlled swap, we further analyze two factors that affect outcomes: the vision pretraining \textit{objective} (classification vs.\ dense objectives such as detection/segmentation), and the vision--language \textit{interface} (input resolution/geometry and connector settings).
Across open-ended VQA and localization benchmarks, we find that SSM-based vision encoders improve localization performance under matched settings while remaining competitive on VQA, and can match or surpass substantially larger backbones on localization and grounding benchmarks.
We further observe that standard vision metrics and naive backbone scaling can misrank VLM performance, and that some \textit{backbone--interface} combinations are sensitive under certain resolution/geometry settings; these findings motivate simple stabilizations and practical guidance for selecting and deploying vision encoders in grounding-sensitive regimes. The overall framework of this paper is shown in Figure~\ref{fig:figure1}.

Our contributions are summarized as follows: 
(i) A controlled evaluation of frozen VLM vision encoders via backbone swaps across Transformer, SSM, and hybrid architectures, together with targeted analyses of pretraining objective and interface choices.
(ii) Empirical evidence that SSM-based vision encoders (VMamba) improve localization under matched settings while remaining competitive on open-ended VQA.
(iii) Systematic experiments reveal overlooked failure modes and show they are fixable 
(e.g., pretraining objective and visual model size do not always correlate with the overall VLM performance).
(iv) We introduce a backbone–objective–interface exploration in the VLM design, and highlight SSM vision encoders as an underexplored, strong alternative.

\section{Preliminaries}
\label{sec:prelim}

\noindent
Our goal is to understand how the choice of vision backbone affects VLM behavior. To attribute differences to the vision encoder rather than to other confounding factors, we keep the rest of the VLM pipeline identical across experiments and swap only the vision backbone checkpoint. Our experimental framework largely follows~\cite{prismatic}; unless we explicitly note a change or highlight a detail for clarity,
we use the same setup. In this section, we summarize the settings shared by all experiments: (1) VLM architecture and notation, (2) optimization, (3) training data and preprocessing, (4) training implementation, and (5) evaluation suite.

\subsection{Model Architecture and Notation.}
\label{sec:vlm_arch}

We adopt a VLM architecture~\cite{llava} consisting of a vision encoder, a lightweight connector, and a decoder-only language model (Vicuna-7B).
A VLM takes an image $x_\text{img} \in \mathbb{R}^{H \times W \times 3}$ and a text prompt $p_\text{prompt}$ in natural language form.

\textit{Vision Encoder.} First, the input image $x_\text{img}$ is fed into vision encoder $V_\omega$ which extract features from the image and output a sequence of tokens $f_\text{img} \in \mathbb{R}^{L \times d_{\text{vision}}}$ where $f_\text{img} = V_\omega(x_\text{img})$, $L$ is the number of visual tokens, and
$d_{\text{vision}}$ is the dimension of a token. $L$ and $d_{\text{vision}}$ are defined by each vision backbones and depend on $H$ and $W$. 

\textit{Connector.} After we get the visual tokens $f_\text{img}$ from the vision encoder, we map these tokens into the LLM embedding space $d_{\text{text}}$ using a connector $C_\psi$. The output visual embeddings are $e_\text{img} \in \mathbb{R}^{L \times d_{\text{text}}}$, where $e_\text{img} = C_\psi(f_\text{img})$.
Unless otherwise specified, the connector is defined as
$C_\psi(x) = W_2 \,\mathrm{GELU}(W_1 x + b_1) + b_2$
where $W_1 \in \mathbb{R}^{d_{\text{text}} \times d_{\text{vis}}}$, $W_2 \in \mathbb{R}^{d_{\text{text}} \times d_{\text{text}}}$, and $b_1, b_2 \in \mathbb{R}^{d_{\text{text}}}$.

\textit{Language Model.}
Let $e_{\text{prompt}} \in \mathbb{R}^{N \times d_{\text{text}}}$ denote the prompt embeddings obtained by applying the LLM's tokenizer and embedding layer to the text prompt $u_{\text{prompt}}$.
We concatenate the visual and text embeddings along the sequence dimension: $\mathbf{e}_{\text{input}} = [e_{\text{img}}; e_{\text{prompt}}] \in \mathbb{R}^{(L+N)\times d_{\text{text}}}.$
The decoder-only language model $f_\theta$ then consumes $\mathbf{e}_{\text{input}}$ and autoregressively generates the output text $u_{\text{gen}} = f_\theta(\mathbf{e}_{\text{input}})$.

\subsection{Optimization.}

\textit{Optimization.}
Prior work finds that one-stage instruction tuning (i.e., jointly training a randomly initialized connector with the LLM) is more efficient and yields better performance than a two-stage pipeline that first aligns the connector and then performs joint tuning~\cite{prismatic}. Following this recipe, we initialize the vision encoder $V_\omega$ and language model $f_\theta$ from pretrained checkpoints, and randomly initialize the connector $C_\psi$. During training, we freeze $\omega$ and update only $(\theta,\psi)$ via instruction tuning.
We fix all optimization hyperparameters (optimizer, learning-rate schedule, batch size, number of steps, and precision) and the random seed across experiments; full details are provided in the Appendix~\ref{app:imp}.

\subsection{Training}
\textit{Data and Preprocessing.}
We fine-tune on 665K multimodal instruction-tuning examples~\cite{prismatic}; a detailed breakdown is provided in the appendix. 

\noindent
For image preprocessing, we apply letterbox resizing, which preserves the original aspect ratio by scaling the image and padding to the target resolution.

\noindent
\textit{Training Implementation.}
We base our training on a verified codebase~\footnote{https://github.com/TRI-ML/prismatic-vlms}, using Fully Sharded Data Parallel (FSDP) on $4\times$ NVIDIA H200 GPUs with a fixed batch order.

\subsection{Evaluation Suite.}
Our evaluation for all the experiments covers two benchmark groups: \textbf{VQA} (VQA-v2~\cite{vqav2}, GQA~\cite{gqa}, VizWiz~\cite{vizwiz}, TextVQA~\cite{textvqa}, POPE~\cite{pope}, TallyQA~\cite{tallyqa}) and \textbf{localization} (RefCOCO, RefCOCO+, RefCOCOg~\cite{refcoco}, OCID-Ref~\cite{ocid}). All pre-/post-processing and dataset-specific thresholds follow~\cite{prismatic}.
We also report the average VQA score, average localization score, and an overall average across all benchmarks weighted by the number of samples in each benchmark, to summarize VQA, localization, and overall performance for each checkpoint.
\section{Investigating Different Vision Encoders}
\label{sec:results}
We report results in two regimes. First, we compare vision encoders under a strictly matched setting to isolate architectural effects in~\ref{sec:results_matched}. Second, we evaluate detection- and segmentation-adapted checkpoints to study the impact of dense objectives in~\ref{sec:results_dense}. Then, we summarize our observations in~\ref{sec:results_summary}. Additional unmatched comparisons against larger data and larger-scale baselines are included in Appendix~\ref{app:unfair_comp}.

\subsection{Matched IN1K/224 Backbone Swaps}
\label{sec:results_matched}
To compare IN1K-pretrained VMamba under a matched backbone-swap setting, we include three representative baselines from distinct architecture families. ViT~\cite{vit} tokenizes an image into fixed-size patches and applies global self-attention over the patch sequence. MaxViT~\cite{maxvit} is a hierarchical hybrid that combines convolutions with multi-axis attention (blocked local and dilated global attention) to capture local and global interactions. MambaVision~\cite{mambavision} is a hybrid Mamba--Transformer backbone that adapts Mamba blocks for vision and retains self-attention in the final layers to capture long-range spatial dependencies. These backbones serve as strong reference points within their respective families.

To enforce a strictly matched setup that isolates backbone architecture, we use checkpoints pretrained on IN1K at 224$\times$224 resolution across all families. For multi-stage backbones (i.e., VMamba, MaxViT, and MambaVision), we extract features from the stage that yields the same number of visual tokens as ViT ($L{=}196$). In Appendix~\ref{app:feature_extract}, we further show that this choice also gives the best performance for these multi-stage backbones among plausible extraction stages.

\begin{table*}[th!] 
    \caption{\textbf{Matched IN1K/224 backbone swaps (VQA).}
    ImageNet-1K supervised vision encoders plugged into the same VLM under a strictly matched setting with 224$\times$224 inputs and $L{=}196$ visual tokens. We report per-benchmark VQA scores and the weighted average VQA score. Across all tables, the best results are shown in bold, and the second best are underlined.
    }
    \vspace{-0.5cm}
    \label{tab:in1k_224-vqa}
      \begin{center}
        \begin{small}
            \resizebox{\textwidth}{!}{%
            \begin{tabular}{c c c | c c c c c c c | c} 
                \toprule
                Visual & Encoder & IN1K & \multirow{2}{*}{VQA-v2} & \multirow{2}{*}{GQA} & \multirow{2}{*}{VizWiz} & TextVQA & \multirow{2}{*}{TextVQA} & \multirow{2}{*}{POPE} & \multirow{2}{*}{TallyQA} & Weighted \\
                Encoder & Size & Acc &  &  &  & +OCR &  &  &  & VQA \\
                \midrule
                ViT-S & 22M & 78.8 & 59.86 & 50.48 & 46.69 & 44.02 & \textbf{12.30} & 77.76 & 48.96 & 57.25\\
                ViT-B & 87M & 81.1 & 59.63 & 49.19 & 45.63 & 44.51 & 12.07 & 75.90 & 50.57 & 57.17\\
                \midrule
                \arrayrulecolor{black}
                MaxViT-T & 31M & 83.41 & 61.08 & 49.43 & 47.58 & 44.31 & 12.02 & 76.77 & 53.85 & 58.75\\
                MaxViT-S & 69M & 84.43 & 59.90 & 48.94 & 41.50 & 44.51 & 11.82 & 76.29 & 51.42 & 57.42\\
                MaxViT-B & 119M & 84.85 & 60.10 & 49.72 & 45.00 & \underline{44.59} & \underline{12.29} & 76.26 & 50.94 & 57.60\\
                MaxViT-L & 212M & 84.93 & 60.07 & 49.55 & 46.42 & 43.67 & 12.23 & 76.46 & 48.86 & 57.30\\
                \midrule
                \arrayrulecolor{black}
                MambaVision-T & 32M & 82.3 & 56.85 & 47.43 & 46.16 & 43.74 & 11.14 & 71.34 & 46.89 & 54.38\\
                MambaVision-T2 & 35M & 82.7 & 56.69 & 47.14 & 43.17 & 43.31 & 10.82 & 69.58 & 43.74 & 53.71\\
                MambaVision-S & 50M & 83.3 & 58.10 & 48.05 & 40.91 & 43.08 & 11.36 & 72.03 & 49.48 & 55.61\\
                MambaVision-B & 98M & 84.2 & 59.14 & 49.23 & 44.21 & 43.33 & 11.43 & 73.05 & 49.53 & 56.53\\
                MambaVision-L & 228M & 85 & 59.26 & 49.56 & 42.13 & 43.75 & 11.41 & 74.46 & 51.70 & 56.94\\
                MambaVision-L2 & 242M & 85.3 & 59.27 & 48.83 & \textbf{48.91} & 43.30 & 11.16 & 74.15 & 50.70 & 56.86\\
                \midrule
                \arrayrulecolor{black}
                VMamba-T & 30M & 82.6 & \underline{64.99} & \underline{54.02} & 44.96 & \textbf{44.62} & 12.22 & \underline{81.70} & \underline{54.58} & \underline{62.07}\\
                VMamba-S & 50M & 83.6 & \textbf{65.24} & \textbf{54.08} & 44.95 & 44.06 & 12.24 & \textbf{82.20} & \textbf{55.54} & \textbf{62.39}\\
                VMamba-B & 80M & 83.9 & 64.20 & 53.25 & \underline{47.97} & 43.81 & 12.08 & 80.75 & 54.05 & 61.38\\
                \bottomrule
            \end{tabular}
            }
    \end{small}
  \end{center}
  \vspace{-0.6cm}
\end{table*}

\begin{table*}[th!] 
    \caption{\textbf{Matched IN1K/224 backbone swaps (localization and overall).}
    Same setting as~\ref{tab:in1k_224-vqa}, reporting per-benchmark localization/grounding scores, the weighted average localization score, and the weighted overall average across all benchmarks.
    }
    \vspace{-0.5cm}
    \label{tab:in1k_224-loc}
      \begin{center}
        \begin{small}
            \resizebox{\textwidth}{!}{%
            \begin{tabular}{c c c | c c c c | c || c} 
                \toprule
                Visual & Encoder & IN1K & \multirow{2}{*}{RefCOCO} & \multirow{2}{*}{RefCOCO+} & \multirow{2}{*}{RefCOCOg} & \multirow{2}{*}{OCID-Ref} & Weighted & Weighted \\
                Encoder & Size & Acc &  &  &  &  & Loc. & Overall \\
                \midrule
                ViT-S & 22M & 78.8 & 32.32 & 21.77 & 24.80 & 5.08 & 17.82 & 51.95\\
                ViT-B & 87M & 81.1 & 26.66 & 15.41 & 19.12 & 4.14 & 13.92 & 51.36\\
                \midrule
                \arrayrulecolor{black}
                MaxViT-T & 31M & 83.41 & 29.44 & 17.29 & 22.10 & 5.17 & 15.79 & 52.98\\
                MaxViT-S & 69M & 84.43 & 25.57 & 14.41 & 17.28 & 4.38 & 13.32 & 51.49\\
                MaxViT-B & 119M & 84.85 & 22.02 & 12.67 & 15.28 & 3.36 & 11.41 & 51.39\\
                MaxViT-L & 212M & 84.93 & 21.15 & 12.01 & 14.77 & 2.94 & 10.81 & 51.05\\
                \midrule
                \arrayrulecolor{black}
                MambaVision-T & 32M & 82.3 & 34.59 & 24.01 & 27.14 & 9.52 & 20.98 & 49.89\\
                MambaVision-T2 & 35M & 82.7 & 35.78 & 24.69 & 27.61 & 9.72 & 21.56 & 49.39\\
                MambaVision-S & 50M & 83.3 & 44.65 & 33.06 & 37.48 & 13.21 & 28.22 & 51.93\\
                MambaVision-B & 98M & 84.2 & 48.52 & 35.57 & 40.58 & 15.84 & 31.17 & 53.12\\
                MambaVision-L & 228M & 85 & 48.19 & 35.82 & 40.26 & 16.79 & 31.51 & 53.52\\
                MambaVision-L2 & 242M & 85.3 & 47.91 & 35.41 & 40.28 & 16.48 & 31.22 & 53.42\\
                \midrule
                \arrayrulecolor{black}
                VMamba-T & 30M & 82.6 & \textbf{58.25} & \textbf{46.64} & \textbf{51.74} & \underline{20.24} & \textbf{39.20} & \underline{59.00}\\
                VMamba-S & 50M & 83.6 & \underline{56.48} & \underline{44.27} & \underline{49.88} & \textbf{23.09} & \underline{39.17} & \textbf{59.27}\\
                VMamba-B & 80M & 83.9 & 42.06 & 31.43 & 36.15 & 15.23 & 27.89 & 56.88\\
                \bottomrule
            \end{tabular}
            }
    \end{small}
  \end{center}
  \vspace{-0.6cm}
\end{table*}

\textit{Observations.}
Under the matched IN1K/224 backbone setting in \autoref{tab:in1k_224-vqa} and \autoref{tab:in1k_224-loc}, VMamba is the strongest across its T/S/B variants, showing better overall performance. 
In addition, VMamba-T/S consistently outperforms other methods on grounding across all localization benchmarks.
For VLMs with ViT and MaxViT backbones, higher IN1K accuracy consistently corresponds to lower VLM performance. In contrast, VLMs with VMamba and MambaVision backbones improve with scaling at small sizes, but show the same degradation at larger scales.

\subsection{Dense Objectives Pretrained Backbone Comparisons}
\label{sec:results_dense}
We next evaluate dense-objective checkpoints with higher resolutions. Alongside VMamba, we include two dense-task baselines: ViTDet~\cite{vitdet}, which adapts a plain ViT backbone for object detection and shows that a simple feature pyramid from a single-scale feature map can suffice for detection fine-tuning, and DeiT~\cite{deit} checkpoints adapted with the ViT-Adapter framework~\cite{vitadapter}, which adds a pre-training-free adapter to a plain ViT to introduce image-specific inductive biases for dense prediction.
Concretely, we use VMamba and ViTDet pretrained on IN1K and then fine-tuned for detection, and VMamba and DeiT pretrained on IN1K and then fine-tuned for segmentation.
Because these checkpoints differ in input geometry and feature extraction stages, they generally produce different output token lengths $L$. We therefore treat these results as evidence about dense pretraining objective effects, rather than as perfectly matched architectural comparisons.\\

\begin{table*}[th!] 
    \caption{\textbf{Dense-objective checkpoints at pretraining resolution (VQA).}
    We compare vision encoders adapted with dense objectives, including detection-pretrained (IN1K$\rightarrow$COCO) ViTDet as a Transformer baseline and VMamba as the SSM backbone, and segmentation-pretrained (IN1K$\rightarrow$ADE20K) DeiT baselines and VMamba. All checkpoints are evaluated at their pretraining input geometries and we report per-benchmark VQA scores and the average VQA score. Because input geometry (and thus token length) differs across entries, these results reflect the impact of the dense pretraining objective rather than perfectly matched architectural swaps.}
    \vspace{-0.4cm}
    \label{tab:detseg-vqa}
      \begin{center}
        \begin{small}
            \resizebox{\textwidth}{!}{%
            \begin{tabular}{c c c c c | c c c c c c c | c} 
                \toprule
                Pretrained & Visual & Encoder & Image & Vision & \multirow{2}{*}{VQA-v2} & \multirow{2}{*}{GQA} & \multirow{2}{*}{VizWiz} & TextVQA & \multirow{2}{*}{TextVQA} & \multirow{2}{*}{POPE} & \multirow{2}{*}{TallyQA} & Weighted \\
                Dataset & Encoder & Size & Size & Token \# &  &  &  & +OCR &  &  &  & VQA \\
                \midrule
                IN1K $\to$ COCO & ViTDet-B & 111M & 1024x1024 & 4096 & \textbf{65.83} & \underline{53.61} & \textbf{50.58} & 43.75 & 11.60 & \underline{84.46} & \underline{55.96} & \textbf{63.00}\\
                IN1K $\to$ COCO & ViTDet-L & 331M & 1024x1024 & 4096 & 60.00 & 48.59 & 44.78 & 43.88 & 11.78 & 78.14 & 51.94 & 57.64\\
                IN1K $\to$ COCO & ViTDet-H & 662M & 1024x1024 & 4096 & 64.43 & 52.62 & 48.99 & 43.93 & 11.50 & 84.29 & 55.95 & 61.89\\
                \cdashline{1-13}
                \arrayrulecolor{black}
                IN1K $\to$ COCO & VMamba-T & 30M & 1333x800 & 4150 & 60.24 & 49.18 & \underline{49.04} & \underline{44.06} & \underline{11.81} & 79.48 & 52.65 & 58.05\\
                IN1K $\to$ COCO & VMamba-S & 50M & 1333x800 & 4150 & \underline{64.87} & \textbf{53.98} & 47.73 & \textbf{44.24} & \textbf{11.93} & \textbf{86.47} & \textbf{59.21} & \underline{62.78}\\
                IN1K $\to$ COCO & VMamba-B & 89M & 1333x800 & 4150 & 60.36 & 49.21 & 46.46 & 43.84 & 11.73 & 81.61 & 55.69 & 58.57\\
                \arrayrulecolor{black}
                \midrule
                \midrule
                \arrayrulecolor{black}
                IN1K $\to$ ADE20K & DeiT-S & 58M & 512x512 & 256 & 61.56 & 52.37 & \textbf{52.16} & 44.25 & 11.74 & 79.82 & 53.60 & 59.36\\
                IN1K $\to$ ADE20K & DeiT-B & 134M & 512x512 & 256 & 63.37 & 53.32 & \underline{49.40} & 43.98 & \underline{11.97} & 81.64 & 53.11 & 60.69\\
                \cdashline{1-13}
                \arrayrulecolor{black}
                IN1K $\to$ ADE20K & VMamba-T & 30M & 512x512 & 1024 & 65.45 & 55.26 & 40.91 & 44.18 & 11.85 & 83.65 & \textbf{55.66} & 62.60\\
                IN1K $\to$ ADE20K & VMamba-S & 50M & 512x512 & 1024 & \textbf{66.42} & \underline{55.68} & 47.44 & \underline{44.42} & 11.86 & \underline{84.01} & \underline{53.91} & \textbf{63.21}\\
                IN1K $\to$ ADE20K & VMamba-B & 89M & 512x512 & 1024 & \underline{66.12} & \textbf{55.89} & 40.19 & \textbf{44.50} & \textbf{12.33} & \textbf{84.39} & 53.61 & \underline{62.87}\\
                \bottomrule
            \end{tabular}
            }
    \end{small}
  \end{center}
\end{table*}

\begin{table*}[th!] 
    \caption{\textbf{Dense-objective checkpoints at pretraining resolution (localization and averages).}
    Same settings as the VQA table, reporting localization/grounding benchmarks, the average localization score, and the overall average across all benchmarks. Checkpoints are evaluated at their pretraining input geometries; since geometry (and token length) differs across entries, the comparisons primarily isolate the effect of dense objectives rather than matched backbone architecture.}
    \vspace{-0.4cm}
    \label{tab:detseg-loc}
      \begin{center}
        \begin{small}
            \resizebox{\textwidth}{!}{%
            \begin{tabular}{c c c c c | c c c c | c || c} 
                \toprule
                Pretrained & Visual & Encoder & Image & Vision & \multirow{2}{*}{RefCOCO} & \multirow{2}{*}{RefCOCO+} & \multirow{2}{*}{RefCOCOg} & \multirow{2}{*}{OCID-Ref} & Weighted & Weighted \\
                Dataset & Encoder & Size & Size & Token \# &  &  &  &  & Loc. & Overall \\
                \midrule
                IN1K $\to$ COCO & ViTDet-B & 111M & 1024x1024 & 4096 & \underline{66.03} & \underline{51.17} & \underline{58.17} & \underline{22.37} & \underline{43.74} & \underline{60.42}\\
                IN1K $\to$ COCO & ViTDet-L & 331M & 1024x1024 & 4096 & 24.62 & 13.79 & 17.44 & 4.62 & 13.05 & 51.65\\
                IN1K $\to$ COCO & ViTDet-H & 662M & 1024x1024 & 4096 & 56.41 & 40.48 & 50.25 & 16.01 & 35.38 & 58.33\\
                \cdashline{1-11}
                \arrayrulecolor{black}
                IN1K $\to$ COCO & VMamba-T & 30M & 1333x800 & 4150 & 29.10 & 16.83 & 19.89 & 3.95 & 14.86 & 52.25\\
                IN1K $\to$ COCO & VMamba-S & 50M & 1333x800 & 4150 & \textbf{69.52} & \textbf{52.87} & \textbf{63.50} & \textbf{28.15} & \textbf{47.94} & \textbf{60.78}\\
                IN1K $\to$ COCO & VMamba-B & 89M & 1333x800 & 4150 & 28.43 & 16.44 & 19.87 & 4.97 & 15.02 & 52.72\\
                \arrayrulecolor{black}
                \midrule
                \midrule
                \arrayrulecolor{black}
                IN1K $\to$ ADE20K & DeiT-S & 58M & 512x512 & 256 & 49.52 & 37.07 & 42.46 & 15.35 & 31.78 & 55.65\\
                IN1K $\to$ ADE20K & DeiT-B & 134M & 512x512 & 256 & 52.17 & 39.78 & 45.61 & 16.22 & 33.77 & 57.07\\
                \cdashline{1-11}
                \arrayrulecolor{black}
                IN1K $\to$ ADE20K & VMamba-T & 30M & 512x512 & 1024 & 62.65 & 51.10 & 57.15 & 24.01 & 43.47 & 60.03\\
                IN1K $\to$ ADE20K & VMamba-S & 50M & 512x512 & 1024 & \textbf{64.17} & \textbf{53.98} & \textbf{59.13} & \underline{24.58} & \underline{44.98} & \textbf{60.76}\\
                IN1K $\to$ ADE20K & VMamba-B & 89M & 512x512 & 1024 & \underline{63.99} & \underline{52.52} & \underline{58.68} & \textbf{25.91} & \textbf{45.08} & \underline{60.48}\\
                \bottomrule
            \end{tabular}
            }
    \end{small}
  \end{center}
\end{table*}

\textit{Observations.}
In ~\autoref{tab:detseg-vqa} and ~\autoref{tab:detseg-loc}, detection adaptation can improve both VQA and localization when fine-tuning remains stable (e.g., ViTDet-B and VMamba-S), but it can also exhibit sharp localization degradation (notably ViTDet-L/H and VMamba-T/B), which we refer to as \emph{localization collapse}. In contrast, segmentation adaptation yields more consistently strong localization across scales: segmentation-adapted VMamba remains strong across sizes and generally outperforms the DeiT (ViT-Adapter) baselines.

\subsection{Summary of Observations}
\label{sec:results_summary}
Across the matched backbone-swap and dense-objective regimes, we observe:
(1) Under strictly matched IN1K/224, $L{=}196$ swaps, VMamba achieves the strongest overall performance, with VMamba-T/S consistently leading localization across all grounding benchmarks, and VMamba variants achieve top aggregate VQA.
(2) Dense pretraining objectives (detection/segmentation) can improve both VQA and localization for SSM- and Transformer-based vision encoders.
(3) ImageNet accuracy and naive backbone scaling
might not directly improve 
downstream VLM performance, and can even degrade.
(4) Some detection-pretrained configurations exhibit sharp localization degradation (\emph{localization collapse}).
Section~\ref{sec:analysis} analyzes these patterns and their causes, and presents practical stabilizations.
\section{Analysis, Diagnosis, and Stabilizations}
\label{sec:analysis}

We first establish the relationship between VQA and localization metrics in~\ref{sec:analysis_loc_importance}, which provides the basis for the analyses in this section. 
We now analyze the empirical patterns from Sec.~\ref{sec:results}: (i) why VMamba-T/S consistently lead localization and why VMamba achieves the strongest overall performance under matched IN1K/224 swaps in Sec.~\ref{sec:analysis_vmamba_loc}; (ii) how dense pretraining objectives (detection/segmentation) affect VQA and localization across backbone families in Sec.~\ref{sec:analysis_dense_objective}; and (iii) two failure modes—ImageNet/scale being an unreliable predictor of downstream VLM behavior, and sharp localization degradation (\emph{localization collapse}) in some high-resolution detection-adapted settings in Sec.~\ref{sec:analysis_failures}. We close with practical implications that motivate the stabilizations in Sec.~\ref{sec:stabilizations}.

\subsection{Localization Matters for General VQA}
\label{sec:analysis_loc_importance}
General open-ended VQA benchmarks implicitly require models to localize relevant objects and regions. For example, GQA is constructed around scene graphs with explicit grounding and evaluates grounding quality, and several works show that improving visual grounding or spatial localization leads to better VQA performance on both GQA and VQA-v2~\cite{gqa,weakly,learning}.\\

To quantify the relationship between VQA and localization in our setting, we compute a Pearson correlation matrix across the metrics reported for all the checkpoint runs. For each run, we take its per-benchmark scores and compute Pearson $r$ between every pair of metrics. The correlation matrix shows that the localization benchmarks are highly consistent with each other, and that VQA-v2, GQA, POPE, and TallyQA correlate moderately to strongly with localization (from 0.65 to 0.80), while TextVQA and VizWiz correlate weakly (from 0.23 to 0.50). The detailed correlation matrix numbers are provided in~\ref{app:corr_matrix}.
Overall, these trends support treating localization as a primary axis when comparing frozen vision encoders, and suggest that differences that are subtle on VQA can be amplified on localization benchmarks.

\begin{figure}[t]
  \centering
  \includegraphics[width=1\linewidth]{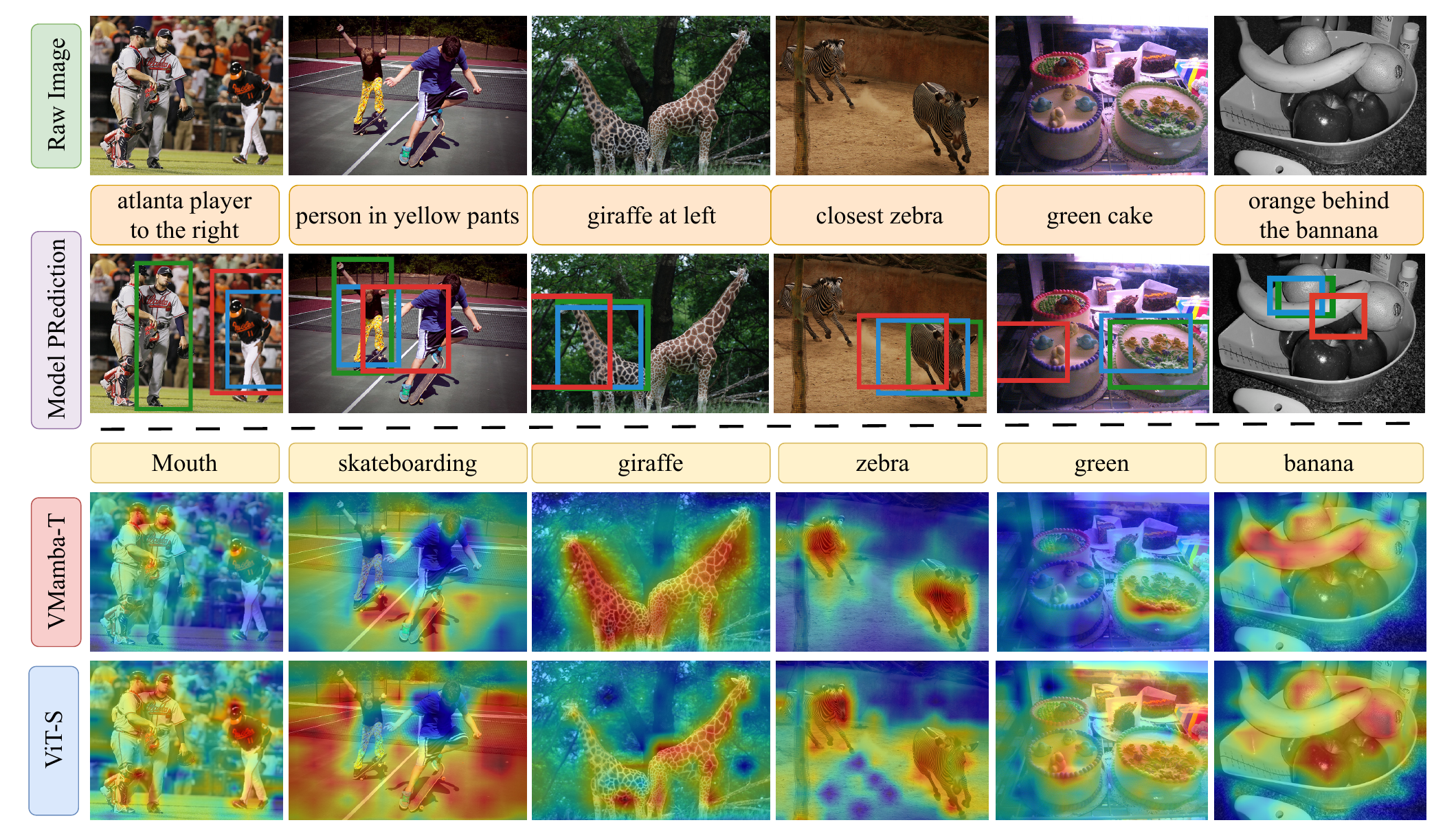}
    \vspace{-0.5cm}
    \caption{
    \textbf{(Top) Grounding examples.} Under the matched IN1K/224 setting, VMamba-T (blue) predicts boxes closer to ground truth (green) than ViT-S (red).
    \textbf{(Bottom) Token--region similarity.} Similarity maps between the visual feature map and the corresponding text tokens from an intermediate LLM layer show VMamba-T yields sharper, more spatially localized text--vision alignment than ViT-S, indicating better preservation and utilization of spatial information.
    }
  \label{fig:similarity}
\end{figure}

\subsection{VMamba is Strong Under the Matched IN1K/224 Setting}
\label{sec:analysis_vmamba_loc}
We next analyze why VMamba performs strongly on localization in the matched IN1K/224 setting. VMamba’s SS4D layer applies state-space aggregation along the 2D token grid in four scan directions; each patch receives four directional state updates that propagate information across rows and columns. This four-directional aggregation is repeated throughout the network, so spatial interactions are baked into the architecture and can preserve spatial structure even under standard classification pretraining without an explicit localization objective. 
In the ViT's case, in contrast, the Transformer is permutation-invariant with respect to token order, so spatial structure is carried mainly by positional encodings. Without an objective that explicitly rewards spatial fidelity, this positional information can be underutilized or attenuated in deeper layers under standard ImageNet pretraining~\cite{droppos,vit_provably}.

To provide qualitative support, we visualize token--region similarity maps (\autoref{fig:similarity}, bottom). In the baseball example (first column), VMamba-T assigns high similarity tightly to the small, discriminative region referenced by the text (Mouth), whereas ViT-S produces a more diffuse response over larger body regions. Across the other examples, VMamba-T consistently yields sharper, more concentrated peaks on the target object, while ViT-S tends to spread high responses across multiple regions and sometimes across multiple objects, suggesting weaker spatial selectivity.

This pattern is consistent with the grounding predictions in \autoref{fig:similarity} (top): VMamba-T’s boxes more closely match the ground-truth regions than ViT-S. Even in the failure case in the first column (selecting the wrong player), VMamba-T still produces a tighter box around the predicted referent, matching the more localized similarity structure observed in the corresponding similarity map. These qualitative results align with the quantitative localization gains in \autoref{tab:in1k_224-loc}.

\subsection{Dense-Task Objectives Help}
\label{sec:analysis_dense_objective}
Since VLM performance depends on spatially grounded visual representations, adapting ImageNet-classification backbones with dense objectives such as detection and segmentation is a natural way to strengthen spatial fidelity. These dense objectives directly supervise spatial layouts and region-level discrimination, and in our experiments this often translates into gains on both localization and VQA, provided the resulting VLM interface remains stable (Sec.~\ref{sec:results_dense}).

This effect is most pronounced for ViT-family baselines that lack built-in spatial inductive bias: detection-adapted ViTDet and segmentation-adapted DeiT substantially improve over their classification-pretrained ViT counterparts. 
VMamba also benefits from dense-objective adaptation, but the improvements are typically more moderate, which is consistent with VMamba already preserving spatial structure under IN1K pretraining through its architectural bias.

Finally, we observe a similar pattern for non-classification pretraining, where combining contrastive-pretrained backbones (e.g., CLIP, SigLIP) with SSL pretrained backbones (e.g., DINOv2) yields the strongest results in our additional experiments, matching the trend reported by prior work~\cite{prismatic}. However, comparable contrastive/SSL SSM-based backbones are not currently available; thus, we treat this direction as out of scope for the main paper and report details in Appendix~\ref{app:unfair_comp}.

\subsection{Diagnosing Failures}
\label{sec:analysis_failures}
We now diagnose two failure modes surfaced in Sec.~\ref{sec:results}. First, ImageNet top-1 accuracy and naive backbone scaling are unreliable predictors for downstream VLM quality. Second, some dense-objective checkpoints exhibit \emph{localization collapse}, where grounding performance drops abruptly after detection-style adaptation.

\subsubsection{Model Size and ImageNet Analysis} 
\label{sec:analysis_misrank}
Table~\ref{tab:in1k_224-vqa} and Table~\ref{tab:in1k_224-loc} contain clear counterexamples where larger models with higher ImageNet accuracy fail to predict VLM performance. MaxViT-L has higher IN1K accuracy than MaxViT-T/S/B but achieves much worse localization. Similarly, scaling from VMamba-T/S to VMamba-B does not improve overall performance.

At first glance, overfitting on the ImageNet might be the reason why larger models reduce transferability at scale; we evaluate linear probing across diverse vision classification datasets and find evidence of degraded transfer for some large supervised models, but there exist models that transfer well and still have poor VLM performance. Thus, probing alone does not fully explain the VLM ranking. We provide the detailed probing results in Appendix~\ref{app:probing}.

Inspired by the dense-tasks pretraining analysis, instead of overfitting on ImageNet-1K, the larger models are more likely to overfit on the \emph{classification objective}, which means that their features only retain the information for classifying the salient object and abandon most of the spatial information. This can explain why even when a visual backbone shows strong accuracy on ImageNet-1K and generalizes well on other classification datasets, it can still perform poorly on the VQA and localization benchmarks.

As observed in \autoref{tab:in1k_224-vqa} and \autoref{tab:in1k_224-loc}, visual backbones with state-space layers, such as VMamba and MambaVision, are more resistant to this \emph{objective overfitting} as their 2D inductive can retain part of the spatial ability. However, as the model size increases, the parameters could still overpower the architectural induction bias, and the localization numbers still drop.

\vspace{-0.25cm}
\subsubsection{Localization Collapse as an Interface Failure Mode}
\label{sec:analysis_collapse}
Some detection-adapted results reveal sharp localization collapses that are difficult to reconcile with naive scaling expectations. For example, ViTDet-L/H collapses despite ViTDet-B being strong, and VMamba-T/B collapse while VMamba-S is strong.

A natural question is whether these failures are simply caused by poor visual feature extraction. However, this explanation is unconvincing: closely related variants in the same backbone family and objective setting can produce strong results, indicating that the underlying vision features can encode rich spatial information. This suggests that collapse is more likely driven by a failure in \emph{vision--language transfer} interface, rather than a complete absence of spatial information in the vision encoder.

We consider two hypotheses. \textbf{(H1) Transmission bottleneck:} the spatial information is present in the features provided by the vision encoder but is not faithfully conveyed through the connector (e.g., the connector capacity is insufficient to preserve the relevant spatial structure when mapping into the LLM embedding space). \textbf{(H2) Utilization bottleneck:} even if the vision backbone encodes rich spatial information, and the connector faithfully conveys the spatial infromation form the vision encoder, the language model may not be able to reliably interpret and use these spatial cues for grounding. Both hypotheses exclude the vision encoder and implicate an interaction between the interface geometry (resolution and aspect ratio), the connector, and the LLM, which motivates the following stabilization strategies.

\vspace{-0.25cm}
\subsection{From Diagnosis to Stabilizations}
\label{sec:stabilizations}

\subsubsection{Stabilization Experiment Protocol}
\label{sec:stab_protocol}
Across all stabilization experiments, we keep the recipe identical as described in Sec.~\ref{sec:prelim}, and change only the factor under study.
We evaluate stabilizations on representative collapse cases (i.e., detection-adapted DetViT-L/H and VMamba-T/B), and we include settings that are not collapsed as controls to verify that stabilizations do not degrade performance when collapse is not present.

Concretely, we test three strategies to separate transmission versus utilization bottlenecks:
(i) increasing connector capacity to improve transmission of spatial structure into the LLM embedding space,
(ii) modifying the interface geometry by changing image resolutions to reduce instability in dense-tasks adapted checkpoints,
and (iii) combining the above two strategies to test whether the effects are complementary.

\begin{table*}[t] 
    \caption{\textbf{Stabilizing localization collapse: VQA benchmarks.}
    We test two interface strategies on the collapse cases: increasing connector capacity with a stronger MLP connector (marked with (f)) and changing the input geometry to a square 512$\times$512 setting while keeping the vision checkpoint fixed. Values in parentheses denote the change relative to the corresponding collapsed variant, which is highlighted in blue.}
    \vspace{-0.5cm}
    \label{tab:stabilizations-vqa}
      \begin{center}
        \begin{small}
            \resizebox{\textwidth}{!}{%
            \begin{tabular}{c c c | c c c c c c c | c} 
                \toprule
                Visual & Encoder & Image & \multirow{2}{*}{VQA-v2} & \multirow{2}{*}{GQA} & \multirow{2}{*}{VizWiz} & TextVQA & \multirow{2}{*}{TextVQA} & \multirow{2}{*}{POPE} & \multirow{2}{*}{TallyQA} & Weighted \\
                Encoder & Size & Size &  &  &  & +OCR &  &  &  & VQA \\
                \midrule
                ViTDet-B & 111M & 1024x1024 & 65.83 & 53.61 & \textbf{50.58} & 43.75 & 11.60 & 84.46 & 55.96 & 63.00\\
                \rowcolor{blue!8}
                ViTDet-L & 331M & 1024x1024 & 60.00 & 48.59 & 44.78 & 43.88 & 11.78 & 78.14 & 51.94 & 57.64\\
                ViTDet-L(f) & 331M & 1024x1024 & 66.43 (\textcolor{green!60!black}{+6.43}) & 53.54 (\textcolor{green!60!black}{+4.95}) & 45.52 (\textcolor{green!60!black}{+0.74}) & 43.76 (\textcolor{red}{-0.12}) & 11.58 (\textcolor{red}{-0.20}) & 84.82 (\textcolor{green!60!black}{+6.68}) & 57.24 (\textcolor{green!60!black}{+5.30}) & 63.55 (\textcolor{green!60!black}{+5.91})\\
                \rowcolor{blue!8}
                ViTDet-H & 662M & 1024x1024 & 64.43 & 52.62 & 48.99 & 43.93 & 11.50 & 84.29 & 55.95 & 61.89\\
                ViTDet-H(f) & 662M & 1024x1024 & \underline{66.89} (\textcolor{green!60!black}{+2.46}) & 54.22 (\textcolor{green!60!black}{+1.60}) & 41.01 (\textcolor{red}{-7.98}) & 44.04 (\textcolor{green!60!black}{+0.11}) & 11.82 (\textcolor{green!60!black}{+0.32}) & 87.03 (\textcolor{green!60!black}{+2.74}) & 58.08 (\textcolor{green!60!black}{+2.13}) & 64.05 (\textcolor{green!60!black}{+2.16})\\
                \midrule
                \arrayrulecolor{black}
                \rowcolor{blue!8}
                VMamba-T & 30M & 1333x800 & 60.24 & 49.18 & \underline{49.04} & 44.06 & 11.81 & 79.48 & 52.65 & 58.05\\
                VMamba-T(f) & 30M & 1333x800 & 64.46 (\textcolor{green!60!black}{+4.22}) & 53.33 (\textcolor{green!60!black}{+4.15}) & 45.21 (\textcolor{red}{-3.83}) & 43.68 (\textcolor{red}{-0.38}) & 11.90 (\textcolor{green!60!black}{+0.09}) & 84.14 (\textcolor{green!60!black}{+4.66}) & 54.27 (\textcolor{green!60!black}{+1.62}) & 61.66 (\textcolor{green!60!black}{+3.61})\\
                VMamba-T & 30M & 512x512 & \textbf{66.91} (\textcolor{green!60!black}{+6.67}) & \underline{55.30} (\textcolor{green!60!black}{+6.12}) & 45.05 (\textcolor{red}{-3.99}) & 44.43 (\textcolor{green!60!black}{+0.37}) & 11.73 (\textcolor{red}{-0.08}) & 84.86 (\textcolor{green!60!black}{+5.38}) & 58.94 (\textcolor{green!60!black}{+6.29}) & \underline{64.22} (\textcolor{green!60!black}{+6.17})\\
                VMamba-T(f) & 30M & 512x512 & 66.77 (\textcolor{green!60!black}{+6.53}) & \textbf{55.41} (\textcolor{green!60!black}{+6.23}) & 42.75 (\textcolor{red}{-6.29}) & 44.28 (\textcolor{green!60!black}{+0.22}) & 11.72 (\textcolor{red}{-0.09}) & 85.71 (\textcolor{green!60!black}{+6.23}) & \underline{60.19} (\textcolor{green!60!black}{+7.54}) & \textbf{64.28} (\textcolor{green!60!black}{+6.23})\\
                VMamba-S & 50M & 1333x800 & 64.87 & 53.98 & 47.73 & 44.24 & 11.93 & 86.47 & 59.21 & 62.78\\
                \rowcolor{blue!8}
                VMamba-B & 89M & 1333x800 & 60.36 & 49.21 & 46.46 & 43.84 & 11.73 & 81.61 & 55.69 & 58.57\\
                VMamba-B(f) & 89M & 1333x800 & 60.85 (\textcolor{green!60!black}{+0.49}) & 50.25 (\textcolor{green!60!black}{+1.04}) & 40.44 (\textcolor{red}{-6.02}) & 43.99 (\textcolor{green!60!black}{+0.15}) & 11.77 (\textcolor{green!60!black}{+0.04}) & 82.71 (\textcolor{green!60!black}{+1.10}) & 55.04 (\textcolor{red}{-0.65}) & 58.84 (\textcolor{green!60!black}{+0.27})\\
                VMamba-B & 89M & 512x512 & 65.70 (\textcolor{green!60!black}{+5.34}) & 55.01 (\textcolor{green!60!black}{+5.80}) & 46.07 (\textcolor{red}{-0.39}) & \underline{44.66} (\textcolor{green!60!black}{+0.82}) & \underline{12.05} (\textcolor{green!60!black}{+0.32}) & \textbf{87.99} (\textcolor{green!60!black}{+6.38}) & 60.07 (\textcolor{green!60!black}{+4.38}) & 63.58 (\textcolor{green!60!black}{+5.01})\\
                VMamba-B(f) & 89M & 512x512 & 65.63 (\textcolor{green!60!black}{+5.27}) & 55.11 (\textcolor{green!60!black}{+5.90}) & 44.13 (\textcolor{red}{-2.33}) & \textbf{44.86} (\textcolor{green!60!black}{+1.02}) & \textbf{12.15} (\textcolor{green!60!black}{+0.42}) & \underline{87.57} (\textcolor{green!60!black}{+5.96}) & \textbf{60.69} (\textcolor{green!60!black}{+5.00}) & 63.58 (\textcolor{green!60!black}{+5.01})\\
                \bottomrule
            \end{tabular}
            }
    \end{small}
  \end{center}
  \vspace{-0.4cm}
\end{table*}

\begin{table*}[t] 
    \caption{\textbf{Stabilizing localization collapse: localization benchmarks.}
    Same settings as Table~\ref{tab:stabilizations-vqa}, reporting grounding metrics and aggregate scores.}
    \vspace{-0.4cm}
    \label{tab:stabilizations-loc}
      \begin{center}
        \begin{small}
            \resizebox{\textwidth}{!}{%
            \begin{tabular}{c c c | c c c c | c || c} 
                \toprule
                Visual & Encoder & Image & \multirow{2}{*}{RefCOCO} & \multirow{2}{*}{RefCOCO+} & \multirow{2}{*}{RefCOCOg} & \multirow{2}{*}{OCID-Ref} & Weighted & Weighted \\
                Encoder & Size & Size &  &  &  &  & Loc. & Overall \\
                \midrule
                ViTDet-B & 111M & 1024x1024 & 66.03 & 51.17 & 58.17 & 22.37 & 43.74 & 60.42\\
                \rowcolor{blue!8}
                ViTDet-L & 331M & 1024x1024 & 24.62 & 13.79 & 17.44 & 4.62 & 13.05 & 51.65\\
                ViTDet-L(f) & 331M & 1024x1024 & 65.73 (\textcolor{green!60!black}{+41.11}) & 52.09 (\textcolor{green!60!black}{+38.30}) & 58.86 (\textcolor{green!60!black}{+41.42}) & 23.87 (\textcolor{green!60!black}{+19.25}) & 44.58 (\textcolor{green!60!black}{+31.53}) & 61.00 (\textcolor{green!60!black}{+9.35})\\
                \rowcolor{blue!8}
                ViTDet-H & 662M & 1024x1024 & 56.41 & 40.48 & 50.25 & 16.01 & 35.38 & 58.33\\
                ViTDet-H(f) & 662M & 1024x1024 & \underline{69.37} (\textcolor{green!60!black}{+12.96}) & \textbf{54.27} (\textcolor{green!60!black}{+13.79}) & \underline{61.97} (\textcolor{green!60!black}{+11.72}) & 26.51 (\textcolor{green!60!black}{+10.50}) & \underline{47.40} (\textcolor{green!60!black}{+12.02}) & \underline{61.81} (\textcolor{green!60!black}{+3.48})\\
                \midrule
                \arrayrulecolor{black}
                \rowcolor{blue!8}
                VMamba-T & 30M & 1333x800 & 29.10 & 16.83 & 19.89 & 3.95 & 14.86 & 52.25\\
                VMamba-T(f) & 30M & 1333x800 & 57.10 (\textcolor{green!60!black}{+28.00}) & 42.55 (\textcolor{green!60!black}{+25.72}) & 50.10 (\textcolor{green!60!black}{+30.21}) & 20.64 (\textcolor{green!60!black}{+16.69}) & 37.93 (\textcolor{green!60!black}{+23.07}) & 58.47 (\textcolor{green!60!black}{+6.22})\\
                VMamba-T & 30M & 512x512 & 61.51 (\textcolor{green!60!black}{+32.41}) & 50.50 (\textcolor{green!60!black}{+33.67}) & 53.76 (\textcolor{green!60!black}{+33.87}) & 28.50 (\textcolor{green!60!black}{+24.55}) & 44.52 (\textcolor{green!60!black}{+29.66}) & 61.57 (\textcolor{green!60!black}{+9.32})\\
                VMamba-T(f) & 30M & 512x512 & 63.69 (\textcolor{green!60!black}{+34.59}) & 52.70 (\textcolor{green!60!black}{+35.87}) & 55.68 (\textcolor{green!60!black}{+35.79}) & \underline{30.87} (\textcolor{green!60!black}{+26.92}) & 46.75 (\textcolor{green!60!black}{+31.89}) & \textbf{61.92} (\textcolor{green!60!black}{+9.67})\\
                VMamba-S & 50M & 1333x800 & \textbf{69.52} & \underline{52.87} & \textbf{63.50} & 28.15 & \textbf{47.94} & 60.78\\
                \rowcolor{blue!8}
                VMamba-B & 89M & 1333x800 & 28.43 & 16.44 & 19.87 & 4.97 & 15.02 & 52.72\\
                VMamba-B(f) & 89M & 1333x800 & 31.47 (\textcolor{green!60!black}{+3.04}) & 19.10 (\textcolor{green!60!black}{+2.66}) & 22.24 (\textcolor{green!60!black}{+2.37}) & 3.95 (\textcolor{red}{-1.02}) & 16.23 (\textcolor{green!60!black}{+1.21}) & 53.11 (\textcolor{green!60!black}{+0.39})\\
                VMamba-B & 89M & 512x512 & 62.77 (\textcolor{green!60!black}{+34.34}) & 50.14 (\textcolor{green!60!black}{+33.70}) & 56.33 (\textcolor{green!60!black}{+36.46}) & 30.00 (\textcolor{green!60!black}{+25.03}) & 45.63 (\textcolor{green!60!black}{+30.61}) & 61.17 (\textcolor{green!60!black}{+8.45})\\
                VMamba-B(f) & 89M & 512x512 & 64.57 (\textcolor{green!60!black}{+36.14}) & 50.99 (\textcolor{green!60!black}{+34.55}) & 57.07 (\textcolor{green!60!black}{+37.20}) & \textbf{31.35} (\textcolor{green!60!black}{+26.38}) & 46.90 (\textcolor{green!60!black}{+31.88}) & 61.34 (\textcolor{green!60!black}{+8.62})\\
                \bottomrule
            \end{tabular}
            }
    \end{small}
  \end{center}
\end{table*}

\subsubsection{Transmission Bottleneck Test}
\label{sec:stab_connector}
We first test whether localization collapse is caused by insufficient transmission capacity at the vision--language interface by increasing connector capacity while keeping all other settings fixed. Specifically, we replace the default 2-layer MLP connector with a 3-layer MLP, using the same initialization policy and training hyperparameters. Results are reported in \autoref{tab:stabilizations-vqa} and \autoref{tab:stabilizations-loc}; models using the stronger connector are marked with (f) in the visual encoder name.

Increasing connector capacity can substantially recover localization in some collapse cases, and the recovered localization typically comes with improved VQA. For example, ViTDet-L collapses under the default connector but recovers with the 3-layer connector, suggesting that the bottleneck lies primarily in the vision--language interface rather than the vision encoder itself. However, a stronger connector does not resolve all failures. For detection-pretrained VMamba-T, the stronger connector improves localization but still falls short of the ImageNet-pretrained VMamba-T baseline. For VMamba-B, the gain is marginal, and the collapse largely persists. These cases indicate that interface instability is not solely due to limited connector expressivity, motivating the geometry-based stabilizations studied next.

\subsubsection{Utilization Bottleneck Test}
\label{sec:stab_square}
We next test a geometry-based stabilization motivated by a consistent difference across VMamba checkpoints: ImageNet-pretrained VMamba (stable) and segmentation-pretrained VMamba (stable) use square inputs, whereas detection-pretrained VMamba uses high-resolution non-square inputs. We hypothesize that the non-square geometry makes it harder for the language model to reliably utilize spatial cues from the visual tokens, contributing to localization collapse.

Keeping the detection-pretrained VMamba-T/B checkpoints and the full training recipe fixed, we evaluate a square input geometry (512$\times$512) and compare against the original non-square setting. The square geometry eliminates the collapse behavior and improves both localization and VQA for detection-pretrained VMamba-T/B, outperforming their ImageNet-pretrained counterparts (\autoref{tab:stabilizations-vqa}, \autoref{tab:stabilizations-loc}). VMamba-S shows a small degradation, which is expected given that 512$\times$512 deviates from its pretraining resolution.

This sensitivity to geometry is consistent with recent findings that VLM behavior can be brittle to input geometry at high resolution~\cite{native,winwin}. We leave a deeper mechanistic analysis of why square geometry improves utilization to future work.

\subsubsection{Applying Connector \& Square Geometry Together}
\label{sec:stab_combo}
Finally, we combine the two strategies above to test whether connector capacity and geometry address complementary parts of the failure.
We apply the stronger connector and square 512x512 resolution settings on VMamba-T and VMamba-B.
\autoref{tab:stabilizations-vqa} and \autoref{tab:stabilizations-loc} show that the combined stabilization yields the greatest and most consistent improvements across the collapse cases.
Empirically, this suggests that both transmission capacity and interface geometry contribute to stable vision--language transfer.

\subsection{Summary: Backbone, Objective, and Interface}
\label{sec:analysis_summary}
Taken together, our results suggest a factorized view where VLM performance is jointly determined by (i) backbone architecture, (ii) pretraining objective, and (iii) the vision--language interface (connector capacity and input geometry). Architectures with stronger spatial inductive bias and dense-task objectives both tend to improve grounding and VQA; however, these gains only translate into downstream VLM quality when the interface can stably transmit and utilize the spatial signal. When the interface is unstable, even strong backbone--objective combinations can exhibit localization collapse and become ineffective.
Notably, factors (ii) and (iii) are largely architecture-agnostic and can benefit across different backbones, while factor (i) remains important for robustness under matched classification pretraining.\\

\section{Related Works}

\textbf{SSMs in Multimodal Models.}
Recent work applies SSMs to vision–language systems primarily on the language and fusion side, where Mamba-style layers replace transformer-based text backbones or multimodal blocks to improve sequence modeling efficiency, while the vision encoder often remains a ViT- or CNN-style backbone (e.g., Cobra~\cite{cobra}, VL-Mamba~\cite{vl_mamba}). More recent work such as CLIMP~\cite{climp} explores fully Mamba-based contrastive vision–language pretraining by replacing both vision and text encoders with SSM architectures, focusing on representation learning rather than downstream generative VLM behavior. These approaches highlight that SSMs can serve as competitive and efficient components for multimodal modeling, but they do not study the effect of different vision backbone architectures under controlled VLM training settings. In parallel, MambaVLT~\cite{mambavlt} uses Mamba blocks for multimodal fusion and evaluates on classification tasks. In contrast, we evaluate SSM vision backbones (VMamba) as frozen vision towers in a generative VLM, isolating their effect on grounding and VQA under matched training conditions.

\section{Conclusion}

In a controlled LLaVA-style setting with frozen vision encoders, we show that SSM-based VMamba is a strong alternative to ViT-family encoders.
First, we show that, under strictly matched IN1K/224 swaps, VMamba variants achieve the strongest overall performances, with VMamba-T/S consistently leading grounding benchmarks, while dense-task pretraining objectives (i.e., detection and segmentation) further improve both VQA and localization across backbone families. At the same time, we find that ImageNet accuracy and naive scaling are unreliable predictors for downstream VLM quality, and that some high-resolution detection-pretrained configurations can suffer sharp localization collapse. We diagnose collapse as a vision--language interface failure mode and show that simple, architecture-agnostic stabilizations such as increasing connector capacity and adjusting interface geometry can be complementary and recover both localization and overall performance. Taken together, the results support a backbone, objective, and interface view of VLM design and suggest that SSM-based vision encoders are a promising, size-efficient backbone choice for VLMs.\\

\noindent
\textbf{Acknowledgments.}\\
This material is based upon work supported by the SUNY AI Platform on Google Cloud (Phase 2). 

\bibliographystyle{splncs04}
\bibliography{main}

\newpage
\clearpage

\appendix

\section{Training Setup and Hyperparameters}
\label{app:imp}

\noindent\textit{Shared training recipe.}
Unless otherwise specified, all models are trained under the same one-stage instruction-tuning recipe to ensure a controlled comparison across vision backbones. Following Sec.~\ref{sec:prelim}, we freeze the vision encoder and train only the language model and connector.

\noindent\textit{Optimization.}
We use Fully Sharded Data Parallel (FSDP) training on 4 GPUs with BF16 mixed precision and activation checkpointing for the LLM. Training is run for 1 epoch with global batch size 128 and per-GPU batch size 16, corresponding to 2 gradient-accumulation steps. We use AdamW with learning rate $2\times10^{-5}$, weight decay $0.1$, max gradient norm $1.0$, and a linear-warmup cosine-decay schedule with warmup ratio $0.03$. These hyperparameters are fixed across all experiments for fairness.

\noindent\textit{Data and preprocessing.}
We fine-tune on the LLaVA-v1.5 665K instruction-tuning mixture described in Sec.~2.3. Images are processed with letterbox resizing, which preserves aspect ratio by resizing and padding to the target resolution. Incomplete batches are retained rather than dropped. For variable-length inputs, text is truncated to the tokenizer or model maximum length, and images are padded within each batch as needed.

\noindent\textit{Implementation note.}
Our implementation is based on a verified training codebase~\footnote{https://github.com/TRI-ML/prismatic-vlms} and uses a fixed batch order across experiments. Additional low-level implementation details are provided in the released code. \textit{The source code will be released upon acceptance.}

\section{Vision Encoder Feature Extraction Stage}
\label{app:feature_extract}

\noindent\textit{Hierarchical backbones.}
VMamba, MaxViT, and MambaVision use a Swin-style four-stage hierarchical architecture. For an input image $x_{\text{img}} \in \mathbb{R}^{H \times W \times 3}$, the first stage has spatial resolution $(H/4)\times(W/4)$, and each subsequent stage downsamples the feature map by a factor of 2 along each spatial dimension. Thus, stage $n \in \{1,2,3,4\}$ has spatial size
\[
H_n \times W_n = \left(\frac{H}{2^{n+1}}\right)\times\left(\frac{W}{2^{n+1}}\right),
\]
with token count
\[
L_n = H_n W_n = \frac{HW}{2^{2(n+1)}}.
\]
Accordingly, the four stages contain
\[
L_1=\frac{HW}{16}, \qquad
L_2=\frac{HW}{64}, \qquad
L_3=\frac{HW}{256}, \qquad
L_4=\frac{HW}{1024}
\]
tokens.

\noindent\textit{Correspondence to ViT features.}
For comparison, a ViT with patch size $P$ produces
\[
L_{\text{ViT}} = \left(\frac{H}{P}\right)\left(\frac{W}{P}\right) = \frac{HW}{P^2}
\]
tokens. For the standard ViT/16 setting ($P=16$), this gives
\[
L_{\text{ViT}}=\frac{HW}{256}=L_3.
\]
Therefore, stage-3 features of these hierarchical backbones have the same token count as ViT/16 features at the same input resolution. For example, at $224\times224$ resolution, stage 3 has spatial size $14\times14$ and hence 196 tokens, exactly matching ViT/16.

\begin{table*}[t] 
    \caption{MaxViT feature-stage ablation on VQA benchmarks. Rows marked (S4) use stage-4 features, while unmarked rows use stage-3 features. Stage 3 matches the ViT/16 token count at the same input resolution, whereas stage 4 uses fewer vision tokens.}
    \vskip -0.1in
    \label{tab:maxvit-stage-comp-vqa}
      \begin{center}
        \begin{small}
            \resizebox{\textwidth}{!}{%
            \begin{tabular}{c c c c c | c c c c c c c | c} 
                \toprule
                Pretrained & Visual & Encoder & Image & Vision & \multirow{2}{*}{VQA-v2} & \multirow{2}{*}{GQA} & \multirow{2}{*}{VizWiz} & TextVQA & \multirow{2}{*}{TextVQA} & \multirow{2}{*}{POPE} & \multirow{2}{*}{TallyQA} & Weighted \\
                Dataset & Encoder & Size & Size & Token \# &  &  &  & +OCR &  &  &  & VQA \\
                \midrule
                \rowcolor{blue!8}
                IN1K & MaxViT-T(S4) & 31M & 224x224 & 49 & 58.36 & 48.75 & \underline{47.29} & 44.03 & \underline{12.23} & 72.87 & 41.22 & 54.89\\
                \rowcolor{blue!8}
                IN1K & MaxViT-S(S4) & 69M & 224x224 & 49 & 59.05 & \underline{49.68} & 41.13 & 44.45 & 12.16 & 75.25 & 49.23 & 56.50\\
                \rowcolor{blue!8}
                IN1K & MaxViT-B(S4) & 119M & 224x224 & 49 & 58.61 & 49.33 & 44.53 & 43.63 & 11.84 & 76.33 & 45.19 & 55.68\\
                \rowcolor{blue!8}
                IN1K & MaxViT-L(S4) & 212M & 224x224 & 49 & 57.66 & 48.86 & 45.95 & 43.91 & 11.93 & 75.34 & 47.76 & 55.29\\
                \arrayrulecolor{gray!80}
                \midrule 
                IN1K & MaxViT-T & 31M & 224x224 & 196 & \textbf{61.08} & 49.43 & \textbf{47.58} & 44.31 & 12.02 & \textbf{76.77} & \textbf{53.85} & \textbf{58.75} (\textcolor{green!60!black}{+3.86})\\
                IN1K & MaxViT-S & 69M & 224x224 & 196 & 59.90 & 48.94 & 41.50 & \underline{44.51} & 11.82 & 76.29 & \underline{51.42} & 57.42 (\textcolor{green!60!black}{+0.92})\\
                IN1K & MaxViT-B & 119M & 224x224 & 196 & \underline{60.10} & \textbf{49.72} & 45.00 & \textbf{44.59} & \textbf{12.29} & 76.26 & 50.94 & \underline{57.60} (\textcolor{green!60!black}{+1.92})\\
                IN1K & MaxViT-L & 212M & 224x224 & 196 & 60.07 & 49.55 & 46.42 & 43.67 & \underline{12.23} & \underline{76.46} & 48.86 & 57.30 (\textcolor{green!60!black}{+2.01})\\
                \bottomrule
            \end{tabular}
            }
    \end{small}
  \end{center}
  \vskip -0.1in            
\end{table*}

\begin{table*}[t] 
    \caption{MaxViT feature-stage ablation on localization benchmarks. Rows marked (S4) use stage-4 features, while unmarked rows use stage-3 features. Using the lower-token stage-4 features substantially degrades grounding and localization performance.}
    \vskip -0.1in
    \label{tab:maxvit-stage-comp-loc}
      \begin{center}
        \begin{small}
            \resizebox{\textwidth}{!}{%
            \begin{tabular}{c c c c c | c c c c | c || c} 
                \toprule
                Pretrained & Visual & Encoder & Image & Vision & \multirow{2}{*}{RefCOCO} & \multirow{2}{*}{RefCOCO+} & \multirow{2}{*}{RefCOCOg} & \multirow{2}{*}{OCID-Ref} & Weighted & Weighted \\
                Dataset & Encoder & Size & Size & Token \# &  &  &  &  & Loc. & Overall \\
                \midrule
                \rowcolor{blue!8}
                IN1K & MaxViT-T(S4) & 31M & 224x224 & 49 & 13.80 & 9.33 & 10.68 & \underline{4.89} & 8.74 & 48.68\\
                \rowcolor{blue!8}
                IN1K & MaxViT-S(S4) & 69M & 224x224 & 49 & 12.63 & 8.60 & 10.07 & 4.27 & 7.96 & 49.97\\
                \rowcolor{blue!8}
                IN1K & MaxViT-B(S4) & 119M & 224x224 & 49 & 12.05 & 8.10 & 9.13 & 4.73 & 7.79 & 49.24\\
                \rowcolor{blue!8}
                IN1K & MaxViT-L(S4) & 212M & 224x224 & 49 & 12.26 & 7.98 & 8.82 & 4.11 & 7.52 & 48.87\\
                \arrayrulecolor{gray!80}
                \midrule 
                IN1K & MaxViT-T & 31M & 224x224 & 196 & \textbf{29.44} & \textbf{17.29} & \textbf{22.10} & \textbf{5.17} & \textbf{15.79} (\textcolor{green!60!black}{+7.05}) & \textbf{52.98} (\textcolor{green!60!black}{+4.30})\\
                IN1K & MaxViT-S & 69M & 224x224 & 196 & \underline{25.57} & \underline{14.41} & \underline{17.28} & 4.38 & \underline{13.32} (\textcolor{green!60!black}{+5.36}) & \underline{51.49} (\textcolor{green!60!black}{+1.52})\\
                IN1K & MaxViT-B & 119M & 224x224 & 196 & 22.02 & 12.67 & 15.28 & 3.36 & 11.41 (\textcolor{green!60!black}{+3.62}) & 51.39 (\textcolor{green!60!black}{+2.15})\\
                IN1K & MaxViT-L & 212M & 224x224 & 196 & 21.15 & 12.01 & 14.77 & 2.94 & 10.81 (\textcolor{green!60!black}{+3.29}) & 51.05 (\textcolor{green!60!black}{+2.18})\\
                \bottomrule
            \end{tabular}
            }
    \end{small}
  \end{center}
  \vskip -0.1in            
\end{table*}
\begin{table*}[t] 
    \caption{MambaVision feature-stage ablation on VQA benchmarks. Rows marked (S4) use stage-4 features, while unmarked rows use stage-3 features. We report both settings to study the trade-off between token count and semantic abstraction.}
    \vskip -0.1in
    \label{tab:mambabision-stage-comp-vqa}
      \begin{center}
        \begin{small}
            \resizebox{\textwidth}{!}{%
            \begin{tabular}{c c c c c | c c c c c c c | c} 
                \toprule
                Pretrained & Visual & Encoder & Image & Vision & \multirow{2}{*}{VQA-v2} & \multirow{2}{*}{GQA} & \multirow{2}{*}{VizWiz} & TextVQA & \multirow{2}{*}{TextVQA} & \multirow{2}{*}{POPE} & \multirow{2}{*}{TallyQA} & Weighted \\
                Dataset & Encoder & Size & Size & Token \# &  &  &  & +OCR &  &  &  & VQA \\
                \midrule
                \rowcolor{blue!8}
                IN1K & MambaVision-T(S4) & 32M & 224x224 & 49 & 60.84 & 51.77 & 45.76 & 43.86 & 12.06 & 79.70 & \textbf{53.51} & 58.68\\
                \rowcolor{blue!8}
                IN1K & MambaVision-T2(S4) & 35M & 224x224 & 49 & 60.85 & 51.26 & 44.02 & 44.11 & \underline{12.54} & 79.18 & 49.56 & 58.11\\
                \rowcolor{blue!8}
                IN1K & MambaVision-S(S4) & 50M & 224x224 & 49 & 61.34 & 51.32 & 42.71 & \textbf{44.52} & 12.35 & 79.77 & \underline{53.49} & 59.00\\
                \rowcolor{blue!8}
                IN1K & MambaVision-B(S4) & 98M & 224x224 & 49 & \textbf{61.87} & \underline{52.16} & 45.17 & 43.43 & \textbf{12.63} & \textbf{80.47} & 52.45 & \textbf{59.34}\\
                \rowcolor{blue!8}
                IN1K & MambaVision-L(S4) & 228M & 224x224 & 49 & 61.52 & \textbf{52.35} & 42.86 & \underline{44.16} & 12.46 & \underline{80.32} & 50.53 & 58.80\\
                \rowcolor{blue!8}
                IN1K & MambaVision-L2(S4) & 242M & 224x224 & 49 & \underline{61.83} & 51.94 & 44.62 & 43.94 & 11.96 & 80.01 & 52.34 & \underline{59.26}\\
                \arrayrulecolor{gray!80}
                \midrule 
                IN1K & MambaVision-T & 32M & 224x224 & 196 & 56.85 & 47.43 & \underline{46.16} & 43.74 & 11.14 & 71.34 & 46.89 & 54.38 (\textcolor{red}{-4.30})\\
                IN1K & MambaVision-T2 & 35M & 224x224 & 196 & 56.69 & 47.14 & 43.17 & 43.31 & 10.82 & 69.58 & 43.74 & 53.71 (\textcolor{red}{-4.40})\\
                IN1K & MambaVision-S & 50M & 224x224 & 196 & 58.10 & 48.05 & 40.91 & 43.08 & 11.36 & 72.03 & 49.48 & 55.61 (\textcolor{red}{-3.39})\\
                IN1K & MambaVision-B & 98M & 224x224 & 196 & 59.14 & 49.23 & 44.21 & 43.33 & 11.43 & 73.05 & 49.53 & 56.53 (\textcolor{red}{-2.81})\\
                IN1K & MambaVision-L & 228M & 224x224 & 196 & 59.26 & 49.56 & 42.13 & 43.75 & 11.41 & 74.46 & 51.70 & 56.94 (\textcolor{red}{-1.86})\\
                IN1K & MambaVision-L2 & 242M & 224x224 & 196 & 59.27 & 48.83 & \textbf{48.91} & 43.30 & 11.16 & 74.15 & 50.70 & 56.86 (\textcolor{red}{-2.40})\\
                \bottomrule
            \end{tabular}
            }
    \end{small}
  \end{center}
  \vskip -0.1in            
\end{table*}

\begin{table*}[t] 
    \caption{MambaVision feature-stage ablation on localization benchmarks. Rows marked (S4) use stage-4 features, while unmarked rows use stage-3 features. Stage 3 provides a more balanced trade-off for localization under a fair token-count comparison to ViT/16-style backbones.}
    \vskip -0.1in
    \label{tab:mambabision-stage-comp-loc}
      \begin{center}
        \begin{small}
            \resizebox{\textwidth}{!}{%
            \begin{tabular}{c c c c c | c c c c | c || c} 
                \toprule
                Pretrained & Visual & Encoder & Image & Vision & \multirow{2}{*}{RefCOCO} & \multirow{2}{*}{RefCOCO+} & \multirow{2}{*}{RefCOCOg} & \multirow{2}{*}{OCID-Ref} & Weighted & Weighted \\
                Dataset & Encoder & Size & Size & Token \# &  &  &  &  & Loc. & Overall \\
                \midrule
                \rowcolor{blue!8}
                IN1K & MambaVision-T(S4) & 32M & 224x224 & 49 & 33.88 & 23.69 & 28.72 & 7.40 & 20.04 & 53.49\\
                \rowcolor{blue!8}
                IN1K & MambaVision-T2(S4) & 35M & 224x224 & 49 & 30.11 & 20.31 & 23.84 & 7.20 & 17.70 & 52.68\\
                \rowcolor{blue!8}
                IN1K & MambaVision-S(S4) & 50M & 224x224 & 49 & 32.87 & 21.65 & 26.06 & 5.95 & 18.42 & 53.55\\
                \rowcolor{blue!8}
                IN1K & MambaVision-B(S4) & 98M & 224x224 & 49 & 36.33 & 25.48 & 31.58 & 9.01 & 22.03 & \textbf{54.33}\\
                \rowcolor{blue!8}
                IN1K & MambaVision-L(S4) & 228M & 224x224 & 49 & 35.34 & 24.22 & 30.37 & 9.35 & 21.50 & \underline{53.79}\\
                \rowcolor{blue!8}
                IN1K & MambaVision-L2(S4) & 242M & 224x224 & 49 & 31.21 & 20.96 & 25.39 & 7.26 & 18.32 & 53.76\\
                \arrayrulecolor{gray!80}
                \midrule 
                IN1K & MambaVision-T & 32M & 224x224 & 196 & 34.59 & 24.01 & 27.14 & 9.52 & 20.98 (\textcolor{green!60!black}{+0.94}) & 49.89 (\textcolor{red}{-3.60})\\
                IN1K & MambaVision-T2 & 35M & 224x224 & 196 & 35.78 & 24.69 & 27.61 & 9.72 & 21.56 (\textcolor{green!60!black}{+3.86}) & 49.39 (\textcolor{red}{-3.29})\\
                IN1K & MambaVision-S & 50M & 224x224 & 196 & 44.65 & 33.06 & 37.48 & 13.21 & 28.22 (\textcolor{green!60!black}{+9.80}) & 51.93 (\textcolor{red}{-1.62})\\
                IN1K & MambaVision-B & 98M & 224x224 & 196 & \textbf{48.52} & \underline{35.57} & \textbf{40.58} & 15.84 & 31.17 (\textcolor{green!60!black}{+9.14}) & 53.12 (\textcolor{red}{-1.21})\\
                IN1K & MambaVision-L & 228M & 224x224 & 196 & \underline{48.19} & \textbf{35.82} & 40.26 & \textbf{16.79} & \textbf{31.51} (\textcolor{green!60!black}{+10.01}) & 53.52 (\textcolor{red}{-0.27})\\
                IN1K & MambaVision-L2 & 242M & 224x224 & 196 & 47.91 & 35.41 & \underline{40.28} & \underline{16.48} & \underline{31.22} (\textcolor{green!60!black}{+12.90}) & 53.42 (\textcolor{red}{-0.34})\\
                \bottomrule
            \end{tabular}
            }
    \end{small}
  \end{center}
  \vskip -0.1in            
\end{table*}

\noindent\textit{Stage choice in our experiments.}
We also report results using stage-4 features for MaxViT and MambaVision. Across Tables~\ref{tab:maxvit-stage-comp-vqa},~\ref{tab:maxvit-stage-comp-loc},~\ref{tab:mambabision-stage-comp-vqa}, and~\ref{tab:mambabision-stage-comp-loc}, we observe that the choice of feature stage has a substantial impact on downstream performance, particularly on localization benchmarks. Notably, stage-4 features often underperform stage-3 features on grounding and localization tasks even when the stage-4 variant uses a higher input resolution and a comparable number of vision tokens, indicating that this effect is not explained by token count alone. A plausible explanation is that stage-4 features are more semantically abstract, whereas stage-3 features retain richer spatial detail. We therefore adopt stage 3 as the main setting, as it both matches the token count of ViT/16 and yields a better balance between VQA and localization performance.

\section{VMamba vs Vim}
\label{app:pure_ssm}

\begin{table*}[t] 
    \caption{Vim vs.\ VMamba on VQA benchmarks under the shared VLM training recipe. VMamba achieves stronger overall performance and is therefore used as the representative SSM-based backbone in the main paper.}

    \vskip -0.1in
    \label{tab:pure-ssm-vqa}
      \begin{center}
        \begin{small}
            \resizebox{\textwidth}{!}{%
            \begin{tabular}{c c c c c | c c c c c c c | c} 
                \toprule
                Pretrained & Visual & Encoder & Image & Vision & \multirow{2}{*}{VQA-v2} & \multirow{2}{*}{GQA} & \multirow{2}{*}{VizWiz} & TextVQA & \multirow{2}{*}{TextVQA} & \multirow{2}{*}{POPE} & \multirow{2}{*}{TallyQA} & Weighted \\
                Dataset & Encoder & Size & Size & Token \# &  &  &  & +OCR &  &  &  & VQA \\
                \midrule
                IN1K & Vim-T & 7M & 224x224 & 196 & 55.83 & 46.93 & 39.39 & 43.82 & 11.67 & 72.91 & 45.69 & 53.40\\
                IN1K & Vim-T-ft & 7M & 224x224 & 729 & 56.82 & 47.62 & 41.51 & 43.56 & 11.76 & 74.69 & 47.73 & 54.52\\
                IN1K & Vim-S & 26M & 224x224 & 196 & 56.84 & 47.73 & 40.47 & 43.81 & 11.58 & 73.58 & 49.58 & 54.74\\
                IN1K & Vim-S-ft & 26M & 224x224 & 729 & 57.68 & 48.55 & \underline{45.68} & 43.78 & 11.89 & 74.31 & 45.96 & 55.02\\
                IN1K & Vim-B & 98M & 224x224 & 196 & 62.93 & 52.11 & 45.52 & \textbf{44.81} & \textbf{12.50} & 78.97 & \underline{55.07} & 60.46\\
                \midrule
                \arrayrulecolor{black}
                IN1K & VMamba-T & 30M & 224x224 & 196 & \underline{64.99} & \underline{54.02} & 44.96 & \underline{44.62} & 12.22 & \underline{81.70} & 54.58 & \underline{62.07}\\
                IN1K & VMamba-S & 50M & 224x224 & 196 & \textbf{65.24} & \textbf{54.08} & 44.95 & 44.06 & \underline{12.24} & \textbf{82.20} & \textbf{55.54} & \textbf{62.39}\\
                IN1K & VMamba-B & 80M & 224x224 & 196 & 64.20 & 53.25 & \textbf{47.97} & 43.81 & 12.08 & 80.75 & 54.05 & 61.38\\
                \bottomrule
            \end{tabular}
            }
    \end{small}
  \end{center}
  \vskip -0.1in            
\end{table*}

\begin{table*}[t] 
    \caption{Vim vs.\ VMamba on localization benchmarks under the shared VLM training recipe. VMamba yields stronger grounding performance overall and is therefore used as the representative SSM-based backbone in the main paper.}

    \vskip -0.1in
    \label{tab:pure-ssm-loc}
      \begin{center}
        \begin{small}
            \resizebox{\textwidth}{!}{%
            \begin{tabular}{c c c c c | c c c c | c || c} 
                \toprule
                Pretrained & Visual & Encoder & Image & Vision & \multirow{2}{*}{RefCOCO} & \multirow{2}{*}{RefCOCO+} & \multirow{2}{*}{RefCOCOg} & \multirow{2}{*}{OCID-Ref} & Weighted & Weighted \\
                Dataset & Encoder & Size & Size & Token \# &  &  &  &  & Loc. & Overall \\
                \midrule
                IN1K & Vim-T & 7M & 224x224 & 196 & 21.35 & 12.42 & 14.99 & 3.51 & 11.21 & 47.73\\
                IN1K & Vim-T-ft & 7M & 224x224 & 729 & 21.28 & 12.30 & 15.13 & 3.96 & 11.37 & 48.72\\
                IN1K & Vim-S & 26M & 224x224 & 196 & 22.33 & 12.60 & 15.32 & 2.70 & 11.20 & 48.88\\
                IN1K & Vim-S-ft & 26M & 224x224 & 729 & 22.99 & 13.09 & 15.01 & 3.57 & 11.80 & 49.21\\
                IN1K & Vim-B & 98M & 224x224 & 196 & 44.62 & 34.10 & 38.54 & 13.17 & 28.56 & 56.17\\
                \midrule
                \arrayrulecolor{black}
                IN1K & VMamba-T & 30M & 224x224 & 196 & \textbf{58.25} & \textbf{46.64} & \textbf{51.74} & \underline{20.24} & \textbf{39.20} & \underline{59.00}\\
                IN1K & VMamba-S & 50M & 224x224 & 196 & \underline{56.48} & \underline{44.27} & \underline{49.88} & \textbf{23.09} & \underline{39.17} & \textbf{59.27}\\
                IN1K & VMamba-B & 80M & 224x224 & 196 & 42.06 & 31.43 & 36.15 & 15.23 & 27.89 & 56.88\\
                \bottomrule
            \end{tabular}
            }
    \end{small}
  \end{center}
  \vskip -0.1in            
\end{table*}

As shown in Table~\ref{tab:pure-ssm-vqa} and Table~~\ref{tab:pure-ssm-loc}, VMamba consistently performs better than Vim in our VLM setting. We therefore adopt VMamba as the representative SSM-based vision backbone in the main paper.

We restrict this comparison to Vim and VMamba because they are among the most established SSM-based vision backbones with mature public implementations. Exploring newer variants built on top of these architectures in the VLM setting is an interesting direction for future work.

\section{Benchmark Correlation Martix}
\label{app:corr_matrix}

\begin{figure}[!htbp]
    \centering
    \includegraphics[width=0.7\linewidth]{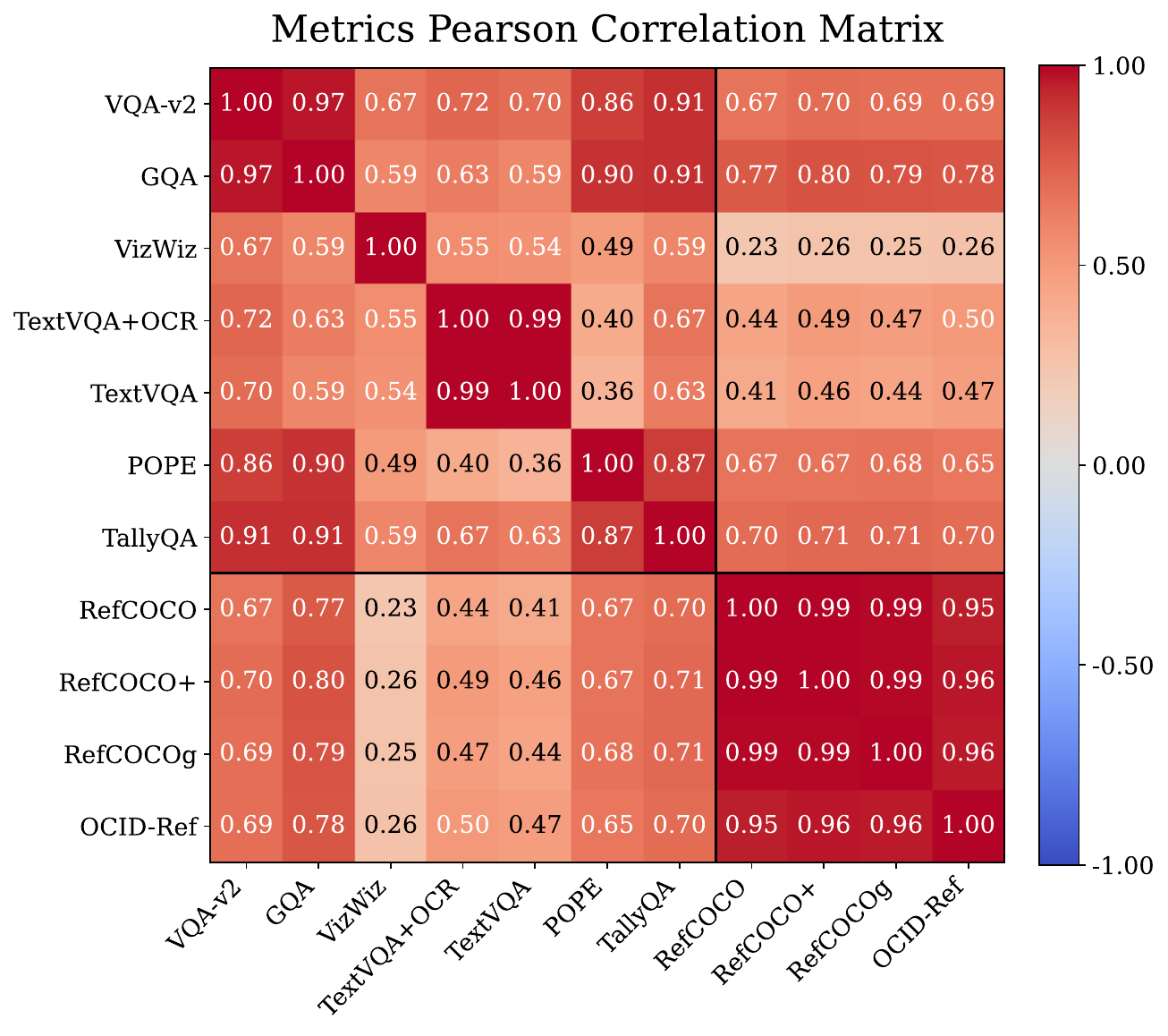}
    \caption{Correlation matrix of benchmark scores across all evaluated models.}
    \label{fig:benchmark_corr}
\end{figure}

Fig.~\ref{fig:benchmark_corr} shows the correlation matrix of benchmark scores across all evaluated models in this study. Each entry represents the Pearson correlation computed using the performance of the same set of models on the corresponding pair of benchmarks. The matrix provides a compact view of how different benchmarks relate to each other across models.

\section{Linear Probing Results}
\label{app:probing}

\begin{table*}[!htbp]
    \caption{Weighted linear probing scores versus downstream weighted VQA, localization, and overall performance on the test split (\%). The Weighted Probing score is the dataset-size-weighted average across probing benchmarks.}
    \vskip -0.1in
    \label{tab:probing-selected-weighted}
      \begin{center}
        \begin{small}
            \resizebox{\textwidth}{!}{
            \begin{tabular}{c c c c | c | c c c}
                \toprule
                Pretrained Dataset & Visual Encoder & Encoder Size & Resolution & Weighted Probing & Weighted VQA & Weighted Loc. & Weighted Overall \\
                \hline
                IN1K & ViT-S & 22M & 224 & 39.19 & 57.25 & 17.82 & 51.95 \\
                IN1K & ViT-B & 87M & 224 & 49.88 & 57.17 & 13.92 & 51.36 \\
                \hline
                IN1K & MaxViT-T & 31M & 224 & 69.22 & 58.75 & 15.79 & 52.98 \\
                IN1K & MaxViT-S & 69M & 224 & 76.23 & 57.42 & 13.32 & 51.49 \\
                IN1K & MaxViT-B & 119M & 224 & 78.13 & 57.60 & 11.41 & 51.39 \\
                IN1K & MaxViT-L & 212M & 224 & 75.92 & 57.30 & 10.81 & 51.05 \\
                \hline
                IN1K & MambaVision-T & 32M & 224 & 37.03 & 54.38 & 20.98 & 49.89 \\
                IN1K & MambaVision-T2 & 35M & 224 & 36.94 & 53.71 & 21.56 & 49.39 \\
                IN1K & MambaVision-S & 50M & 224 & 41.74 & 55.61 & 28.22 & 51.93 \\
                IN1K & MambaVision-B & 98M & 224 & 43.20 & 56.53 & 31.17 & 53.12 \\
                IN1K & MambaVision-L & 228M & 224 & 49.31 & 56.94 & 31.51 & 53.52 \\
                IN1K & MambaVision-L2 & 242M & 224 & 48.32 & 56.86 & 31.22 & 53.42 \\
                \hline
                IN1K & VMamba-T & 30M & 224 & 73.73 & 62.07 & 39.20 & 59.00 \\
                IN1K & VMamba-S & 50M & 224 & \textbf{79.45} & 62.39 & 39.17 & 59.27 \\
                IN1K & VMamba-B & 80M & 224 & \underline{78.88} & 61.38 & 27.89 & 56.88 \\
                \arrayrulecolor{black}
                \hline
                \hline
                \arrayrulecolor{black}
                IN1K $\to$ COCO & ViTDet-B & 111M & 1024 & 52.78 & \underline{63.00} & 43.74 & 60.42 \\
                IN1K $\to$ COCO & ViTDet-L & 331M & 1024 & 56.23 & 57.64 & 13.05 & 51.65 \\
                IN1K $\to$ COCO & ViTDet-H & 662M & 1024 & 54.54 & 61.89 & 35.38 & 58.33 \\
                \hline
                IN1K $\to$ COCO & VMamba-T & 30M & 800 & 50.21 & 58.05 & 14.86 & 52.25 \\
                IN1K $\to$ COCO & VMamba-S & 50M & 800 & 56.54 & 62.78 & \textbf{47.94} & \textbf{60.78} \\
                IN1K $\to$ COCO & VMamba-B & 89M & 800 & 58.24 & 58.57 & 15.02 & 52.72 \\
                \hline
                IN1K $\to$ ADE20K & DeiT-S & 58M & 512 & 54.49 & 59.36 & 31.78 & 55.65 \\
                IN1K $\to$ ADE20K & DeiT-B & 134M & 512 & 67.08 & 60.69 & 33.77 & 57.07 \\
                \hline
                IN1K $\to$ ADE20K & VMamba-T & 30M & 512 & 58.15 & 62.60 & 43.47 & 60.03 \\
                IN1K $\to$ ADE20K & VMamba-S & 50M & 512 & 62.35 & \textbf{63.21} & 44.98 & \underline{60.76} \\
                IN1K $\to$ ADE20K & VMamba-B & 89M & 512 & 65.32 & 62.87 & \underline{45.08} & 60.48 \\
                \bottomrule
            \end{tabular}
            }
    \end{small}
  \end{center}
  \vskip -0.1in
\end{table*}

\begin{table*}[!htbp]
    \caption{Per-dataset linear probing results on the test split (\%). Results are obtained by training a linear classifier on frozen global image features.}

    \vskip -0.1in
    \label{tab:probing-selected-datasets}
      \begin{center}
        \begin{small}
            \resizebox{\textwidth}{!}{
            \begin{tabular}{c | c c c c c c c c c c c c c}
                \toprule
                Visual & \multirow{2}{*}{food101} & \multirow{2}{*}{cifar10} & \multirow{2}{*}{cifar100} & \multirow{2}{*}{dtd} & oxford & stanford & fgvc & \multirow{2}{*}{sun397} & oxford & \multirow{2}{*}{caltech101} & \multirow{2}{*}{birdsnap} & \multirow{2}{*}{voc2007} & coco \\
                Encoder &  &  &  &  & pets & cars & aircraft &  & flowers102 &  &  &  & multilabel \\
                \hline
                ViT-S & 23.56 & 78.83 & 51.55 & 53.88 & 25.02 & 7.20 & 8.94 & 45.15 & 28.90 & 71.10 & 37.09 & 36.69 & 42.60 \\
                ViT-B & 39.58 & 87.88 & 62.13 & 54.31 & 60.53 & 9.75 & 16.26 & 49.43 & 30.54 & 77.61 & 68.83 & 54.92 & 50.87 \\
                \hline
                MaxViT-T & 71.64 & 92.23 & 76.40 & 75.11 & 74.76 & 30.53 & 28.86 & 68.69 & 77.69 & 89.99 & 59.35 & 61.57 & 61.38 \\
                MaxViT-S & 76.91 & 94.82 & 80.56 & \textbf{77.87} & 86.64 & 46.61 & \underline{37.77} & 71.43 & \textbf{86.27} & 92.62 & 88.61 & 78.69 & 68.13 \\
                MaxViT-B & 77.71 & \textbf{96.51} & 82.21 & 73.99 & 92.31 & \underline{54.46} & 36.51 & 70.91 & 84.75 & 92.95 & \underline{98.14} & \underline{86.11} & 73.63 \\
                MaxViT-L & 75.72 & 95.87 & 80.82 & 69.41 & 92.29 & 49.94 & 34.92 & 68.92 & 73.80 & 90.65 & \textbf{98.90} & 84.98 & 74.51 \\
                \hline
                MambaVision-T & 38.08 & 63.25 & 35.93 & 52.82 & 18.34 & 8.32 & 7.68 & 41.41 & 43.32 & 62.82 & 11.42 & 21.74 & 32.27 \\
                MambaVision-T2 & 37.43 & 62.94 & 35.96 & 52.87 & 18.70 & 8.63 & 8.22 & 41.14 & 43.84 & 63.17 & 11.39 & 22.58 & 32.36 \\
                MambaVision-S & 43.30 & 67.77 & 42.14 & 59.41 & 22.19 & 10.15 & 9.78 & 46.59 & 48.98 & 67.44 & 15.26 & 25.69 & 36.56 \\
                MambaVision-B & 45.58 & 69.28 & 43.97 & 59.57 & 22.27 & 10.48 & 9.15 & 48.28 & 51.72 & 67.97 & 16.34 & 25.71 & 37.90 \\
                MambaVision-L & 51.68 & 73.85 & 50.26 & 65.00 & 31.43 & 17.07 & 13.23 & 53.45 & 60.11 & 74.52 & 23.34 & 35.16 & 42.22 \\
                MambaVision-L2 & 51.11 & 72.46 & 48.57 & 64.26 & 29.11 & 16.08 & 12.96 & 52.77 & 58.56 & 73.44 & 22.11 & 33.73 & 41.75 \\
                \hline
                VMamba-T & 76.09 & 94.29 & 80.12 & \underline{76.01} & 88.14 & 32.22 & 30.18 & 71.85 & 81.12 & 92.08 & 79.10 & 70.26 & 68.11 \\
                VMamba-S & \textbf{80.38} & \underline{96.13} & \textbf{83.65} & 75.96 & \underline{93.51} & \textbf{55.60} & \textbf{39.00} & \textbf{74.63} & \underline{85.54} & \textbf{93.11} & 89.79 & 83.84 & 73.56 \\
                VMamba-B & \underline{80.13} & 95.96 & \underline{82.25} & 75.27 & \textbf{94.06} & 53.51 & 32.55 & \underline{74.14} & 80.94 & \underline{93.10} & 90.54 & \textbf{88.04} & 76.01 \\
                \arrayrulecolor{black}
                \hline
                \hline
                \arrayrulecolor{black}
                ViTDet-B & 59.25 & 66.14 & 27.26 & 59.89 & 36.17 & 10.01 & 18.12 & 59.07 & 53.37 & 76.92 & 43.64 & 74.81 & 74.15 \\
                ViTDet-L & 66.78 & 69.22 & 26.38 & 65.16 & 38.51 & 10.73 & 20.82 & 61.37 & 54.37 & 82.08 & 43.66 & 79.46 & 78.86 \\
                ViTDet-H & 66.54 & 74.12 & 29.20 & 62.98 & 26.55 & 10.10 & 16.44 & 57.83 & 44.59 & 77.70 & 35.88 & 78.54 & \underline{80.89} \\
                \hline
                VMamba-T & 59.94 & 53.57 & 24.11 & 58.78 & 39.96 & 10.51 & 18.72 & 55.07 & 60.90 & 76.33 & 41.96 & 57.58 & 69.09 \\
                VMamba-S & 69.50 & 52.74 & 22.55 & 60.80 & 59.63 & 14.14 & 21.18 & 62.10 & 63.12 & 81.67 & 47.18 & 76.01 & 80.10 \\
                VMamba-B & 69.60 & 57.08 & 28.14 & 61.22 & 56.39 & 13.07 & 22.32 & 63.30 & 66.01 & 82.07 & 52.07 & 79.43 & \textbf{83.84} \\
                \hline
                DeiT-S & 63.56 & 50.26 & 22.22 & 65.21 & 63.89 & 13.69 & 14.64 & 68.33 & 56.07 & 79.83 & 56.51 & 61.49 & 61.56 \\
                DeiT-B & 76.31 & 61.22 & 34.74 & 71.54 & 88.44 & 29.81 & 22.32 & 73.68 & 76.47 & 88.18 & 89.06 & 80.97 & 72.03 \\
                \hline
                VMamba-T & 67.47 & 60.79 & 30.58 & 69.79 & 70.37 & 13.92 & 14.55 & 67.62 & 66.95 & 77.71 & 63.01 & 61.17 & 65.16 \\
                VMamba-S & 75.26 & 57.21 & 24.41 & 71.28 & 85.39 & 20.51 & 16.20 & 70.46 & 68.89 & 84.22 & 79.90 & 68.13 & 71.67 \\
                VMamba-B & 76.40 & 63.21 & 31.82 & 72.13 & 88.50 & 21.07 & 19.98 & 72.72 & 68.55 & 86.59 & 82.53 & 76.90 & 75.34 \\
                \bottomrule
            \end{tabular}
            }
    \end{small}
  \end{center}
  \vskip -0.1in
\end{table*}

We evaluate the representation quality of the vision backbones using standard linear probing. For each backbone, we first extract and cache global image features from the frozen encoder for the train, validation, and test splits of each dataset. For each image, we extract a single global feature from the frozen encoder: we use the CLS token for backbones that provide one, and otherwise apply mean pooling over spatial tokens.
A linear classifier (single fully connected layer) is then trained on top of the frozen features. We train the probe for 30 epochs using AdamW with learning rate $10^{-3}$ and batch size 512. Model selection is performed using the validation split, with weight decay selected from a grid $\{10^{-6}, \ldots, 10^{-1}\}$. The best model according to the validation metric is evaluated on the test split.
For multiclass datasets, we report top-1 accuracy, while multilabel datasets use mean average precision (mAP). Table~\ref{tab:probing-selected-weighted} and Table~\ref{tab:probing-selected-datasets} report the resulting test performance. The \textit{Weighted Avg} column is computed as the dataset-size-weighted average across all available benchmarks.

\noindent\textit{Observations.}
Tables~\ref{tab:probing-selected-weighted} and~\ref{tab:probing-selected-datasets} show that linear probing only partially explains downstream VLM behavior. In some families, larger supervised models exhibit slightly worse probing performance, such as MaxViT-L relative to MaxViT-B, MambaVision-L2 relative to MambaVision-L, and VMamba-B relative to VMamba-S, which is consistent with reduced transferability at scale. However, higher probing performance still does not reliably predict stronger VLM performance: ViT-B probes better than ViT-S yet performs worse overall, and MaxViT-S/B/L all probe substantially better than MaxViT-T while yielding weaker localization and lower overall VLM performance. This supports the interpretation in Sec.~\ref{sec:analysis_failures} that the core issue is not merely overfitting to ImageNet-1K, but over-specialization to classification objectives, which favors global category information over spatially grounded representations. Conversely, dense-objective checkpoints can improve downstream VLM performance despite only modest, or even lower, probing scores relative to their classification-pretrained counterparts, indicating that standard global-feature probing does not fully capture the spatial fidelity that matters for grounded VLM behavior.

\section{Unfair Comparisons \& Contrastive/Self-Supervised Backbones}
\label{app:unfair_comp}

\subsection{Unfair Comparisons}
We additionally report several comparisons that are informative but fall outside the controlled setting of the main paper. In these experiments, multiple factors change at once, including the pretraining dataset, input resolution, number of visual tokens, and model size. As a result, these tables should be interpreted as supplementary reference rather than evidence for the main controlled conclusions.

\noindent\textit{Different classification pretraining settings.}
Tables~\ref{tab:cls_unfair_pretrain-vqa},~\ref{tab:cls_unfair_pretrain-loc},~\ref{tab:cls_unfair_res_pretrain-vqa}, and~\ref{tab:cls_unfair_res_pretrain-loc} report classification-pretrained checkpoints that differ from the available VMamba setting in pretraining data scale and/or evaluation resolution. 
ViT and several MaxViT variants generally benefit on VQA from larger-scale classification pretraining or higher-resolution feature extraction, but these changes do not consistently improve localization.
MambaVision does not show the same trend: increasing pretraining scale or resolution does not reliably improve either VQA or localization, and can even degrade performance relative to the smaller controlled setting. In contrast, VMamba remains competitive on VQA and continues to dominate localization despite using a much smaller classification pretraining dataset, lower input resolution, and smaller model sizes. This further supports the summary in Sec.~\ref{sec:analysis} that the inductive bias is important when the pretraining objective is not encouraging the backbone to preserve the spatial information.

\begin{table*}[!htbp]
    \caption{Representative VQA comparison at matched token count. All models are evaluated in the standard $224\times224$ / 196-token setting, or the stage-3 equivalent for hierarchical backbones, providing a compact subset of the main comparison.}

    \vskip -0.1in
    \label{tab:cls_unfair_pretrain-vqa}
      \begin{center}
        \begin{small}
            \resizebox{\textwidth}{!}{
            \begin{tabular}{c c c c c | c c c c c c c | c} 
                \toprule
                Pretrained & Visual & Encoder & Image & Vision & \multirow{2}{*}{VQA-v2} & \multirow{2}{*}{GQA} & \multirow{2}{*}{VizWiz} & TextVQA & \multirow{2}{*}{TextVQA} & \multirow{2}{*}{POPE} & \multirow{2}{*}{TallyQA} & Weighted \\
                Dataset & Encoder & Size & Size & Token \# &  &  &  & +OCR &  &  &  & VQA \\
                \midrule
                IN21k & ViT-T & 5.7M & 224x224 & 196 & 62.82 & 51.30 & 46.03 & \underline{44.62} & 11.90 & 76.94 & 52.65 & 59.95\\
                IN21k & ViT-S & 22M & 224x224 & 196 & 61.28 & 50.34 & 47.95 & 44.18 & 12.30 & 78.59 & 53.62 & 58.97\\
                IN21k & ViT-B & 87M & 224x224 & 196 & \textbf{68.16} & \textbf{55.96} & \underline{49.40} & \textbf{44.75} & \textbf{12.77} & \underline{82.34} & 55.26 & \textbf{64.70}\\
                IN21k & ViT-L & 304M & 224x224 & 196 & 67.13 & 54.85 & 49.15 & 44.28 & 12.64 & \textbf{82.38} & 53.32 & 63.61\\
                \cdashline{1-13}
                \arrayrulecolor{black}
                IN21k $\to$ IN1K & ViT-T & 5.7M & 224x224 & 196 & 61.34 & 50.00 & 45.30 & 44.02 & 11.84 & 77.90 & 53.82 & 58.96\\
                IN21k $\to$ IN1K & ViT-S & 22M & 224x224 & 196 & 60.39 & 49.46 & 44.54 & 44.58 & 12.23 & 78.49 & 51.60 & 57.95\\
                IN21k $\to$ IN1K & ViT-B & 87M & 224x224 & 196 & \underline{67.24} & \underline{54.96} & 47.76 & 43.92 & 12.34 & 82.16 & \textbf{57.16} & \underline{64.17}\\
                IN21k $\to$ IN1K & ViT-L & 304M & 224x224 & 196 & 62.17 & 50.82 & 42.10 & 44.40 & \underline{12.71} & 78.42 & 53.47 & 59.55\\
                \midrule
                \arrayrulecolor{black}
                IN21K & MaxViT-B & 119M & 224x224 & 196 & 62.67 & 49.98 & 45.92 & 44.08 & 12.38 & 76.46 & 53.58 & 59.88\\
                IN21K & MaxViT-L & 212M & 224x224 & 196 & 65.50 & 52.03 & 47.31 & 44.30 & 12.49 & 80.59 & 53.62 & 62.23\\
                IN21K & MaxViT-XL & 475M & 224x224 & 196 & 65.55 & 51.61 & \textbf{50.14} & 44.13 & 12.58 & 80.07 & 54.78 & 62.43\\
                \midrule
                \arrayrulecolor{black}
                IN21K & MambaVision-B & 98M & 224x224 & 196 & 49.42 & 42.07 & 41.98 & 42.71 & 10.36 & 58.90 & 42.05 & 47.50\\
                IN21K & MambaVision-L & 228M & 224x224 & 196 & 52.06 & 43.27 & 39.10 & 43.12 & 10.64 & 65.22 & 43.49 & 49.87\\
                \midrule
                \arrayrulecolor{black}
                IN1K & VMamba-T & 30M & 224x224 & 196 & 64.99 & 54.02 & 44.96 & \underline{44.62} & 12.22 & 81.70 & 54.58 & 62.07\\
                IN1K & VMamba-S & 50M & 224x224 & 196 & 65.24 & 54.08 & 44.95 & 44.06 & 12.24 & 82.20 & \underline{55.54} & 62.39\\
                IN1K & VMamba-B & 80M & 224x224 & 196 & 64.20 & 53.25 & 47.97 & 43.81 & 12.08 & 80.75 & 54.05 & 61.38\\
                \bottomrule
            \end{tabular}
            }
    \end{small}
  \end{center}
  \vskip -0.1in            
\end{table*}

\begin{table*}[!htbp]
    \caption{Representative localization comparison at matched token count. All models are evaluated in the standard $224 \times 224$ / 196-token setting, or the stage-3 equivalent for hierarchical backbones, providing a compact subset of the main comparison.}

    \vskip -0.1in
    \label{tab:cls_unfair_pretrain-loc}
      \begin{center}
        \begin{small}
            \resizebox{\textwidth}{!}{%
            \begin{tabular}{c c c c c | c c c c | c || c} 
                \toprule
                Pretrained & Visual & Encoder & Image & Vision & \multirow{2}{*}{RefCOCO} & \multirow{2}{*}{RefCOCO+} & \multirow{2}{*}{RefCOCOg} & \multirow{2}{*}{OCID-Ref} & Weighted & Weighted \\
                Dataset & Encoder & Size & Size & Token \# &  &  &  &  & Loc. & Overall \\
                \midrule
                IN21k & ViT-T & 5.7M & 224x224 & 196 & 34.52 & 21.72 & 26.47 & 7.30 & 19.43 & 54.50\\
                IN21k & ViT-S & 22M & 224x224 & 196 & 24.40 & 13.79 & 17.42 & 4.73 & 13.04 & 52.80\\
                IN21k & ViT-B & 87M & 224x224 & 196 & 55.00 & 44.79 & 48.35 & 19.90 & 37.46 & \textbf{61.04}\\
                IN21k & ViT-L & 304M & 224x224 & 196 & 40.64 & 31.38 & 35.21 & 11.40 & 25.86 & 58.54\\
                \cdashline{1-11}
                \arrayrulecolor{black}
                IN21k $\to$ IN1K & ViT-T & 5.7M & 224x224 & 196 & 31.56 & 19.84 & 23.79 & 8.35 & 18.40 & 53.51\\
                IN21k $\to$ IN1K & ViT-S & 22M & 224x224 & 196 & 24.29 & 14.05 & 16.99 & 4.05 & 12.75 & 51.88\\
                IN21k $\to$ IN1K & ViT-B & 87M & 224x224 & 196 & 54.82 & \underline{45.00} & 48.30 & 18.44 & 36.87 & \underline{60.50}\\
                IN21k $\to$ IN1K & ViT-L & 304M & 224x224 & 196 & 22.92 & 13.04 & 15.77 & 3.18 & 11.69 & 53.12\\
                \midrule
                \arrayrulecolor{black}
                IN21K & MaxViT-B & 119M & 224x224 & 196 & 23.04 & 12.72 & 16.44 & 2.56 & 11.46 & 53.38\\
                IN21K & MaxViT-L & 212M & 224x224 & 196 & 26.27 & 15.83 & 19.53 & 4.94 & 14.30 & 55.79\\
                IN21K & MaxViT-XL & 475M & 224x224 & 196 & 24.43 & 13.70 & 18.34 & 4.46 & 13.02 & 55.79\\
                \midrule
                \arrayrulecolor{black}
                IN21K & MambaVision-B & 98M & 224x224 & 196 & 19.44 & 10.68 & 12.99 & 3.52 & 10.12 & 42.48\\
                IN21K & MambaVision-L & 228M & 224x224 & 196 & 22.52 & 11.91 & 14.69 & 3.91 & 11.50 & 44.72\\
                \midrule
                \arrayrulecolor{black}
                IN1K & VMamba-T & 30M & 224x224 & 196 & \textbf{58.25} & \textbf{46.64} & \textbf{51.74} & \underline{20.24} & \textbf{39.20} & 59.00\\
                IN1K & VMamba-S & 50M & 224x224 & 196 & \underline{56.48} & 44.27 & \underline{49.88} & \textbf{23.09} & \underline{39.17} & 59.27\\
                IN1K & VMamba-B & 80M & 224x224 & 196 & 42.06 & 31.43 & 36.15 & 15.23 & 27.89 & 56.88\\
                \bottomrule
            \end{tabular}
            }
    \end{small}
  \end{center}
  \vskip -0.1in            
\end{table*}

\begin{table*}[!htbp] 
    \caption{Additional VQA results for selected MaxViT and MambaVision configurations beyond the matched 196-token setting. We report larger-resolution and larger-token variants for hierarchical backbones and compare them with the standard VMamba models.}

    \vskip -0.1in
    \label{tab:cls_unfair_res_pretrain-vqa}
      \begin{center}
        \begin{small}
            \resizebox{\textwidth}{!}{%
            \begin{tabular}{c c c c c | c c c c c c c | c} 
                \toprule
                Pretrained & Visual & Encoder & Image & Vision & \multirow{2}{*}{VQA-v2} & \multirow{2}{*}{GQA} & \multirow{2}{*}{VizWiz} & TextVQA & \multirow{2}{*}{TextVQA} & \multirow{2}{*}{POPE} & \multirow{2}{*}{TallyQA} & Weighted \\
                Dataset & Encoder & Size & Size & Token \# &  &  &  & +OCR &  &  &  & VQA \\
                \midrule
                IN1K & MaxViT-T & 31M & 384x384 & 576 & 64.04 & 53.04 & 48.00 & \underline{44.69} & 11.71 & 79.86 & 53.30 & 61.13\\
                IN1K & MaxViT-S & 69M & 384x384 & 576 & 61.31 & 49.85 & 46.73 & 44.19 & 11.79 & 77.57 & 52.05 & 58.70\\
                IN1K & MaxViT-B & 119M & 384x384 & 576 & 63.90 & 52.76 & 46.61 & 43.95 & 12.27 & 81.23 & 52.83 & 60.97\\
                IN1K & MaxViT-L & 212M & 384x384 & 576 & 62.90 & 51.26 & 49.94 & 43.80 & 12.41 & 81.00 & 49.25 & 59.73\\
                IN1K & MaxViT-T & 31M & 512x512 & 1024 & 64.25 & 52.84 & 48.14 & 44.40 & 11.84 & 79.24 & 51.93 & 61.08\\
                IN1K & MaxViT-S & 69M & 512x512 & 1024 & 63.51 & 51.26 & 46.29 & 44.25 & 11.92 & 79.36 & 54.01 & 60.71\\
                IN1K & MaxViT-B & 119M & 512x512 & 1024 & 64.53 & 53.14 & 47.92 & 44.10 & 12.09 & 82.90 & 53.06 & 61.56\\
                IN1K & MaxViT-L & 212M & 512x512 & 1024 & 63.19 & 51.17 & 46.09 & 43.63 & 12.41 & 81.30 & 48.87 & 59.84\\
                \cdashline{1-13}
                \arrayrulecolor{black}
                IN21K $\to$ IN1K & MaxViT-B & 119M & 384x384 & 576 & 63.44 & 50.92 & 50.41 & 44.38 & 12.71 & 76.72 & 54.53 & 60.71\\
                IN21K $\to$ IN1K & MaxViT-L & 212M & 384x384 & 576 & 67.15 & 52.82 & \textbf{53.89} & 43.72 & \textbf{13.28} & 82.30 & \underline{55.45} & \textbf{63.89}\\
                IN21K $\to$ IN1K & MaxViT-XL & 475M & 384x384 & 576 & 66.37 & 52.53 & 49.11 & 44.16 & 12.89 & 81.46 & 53.58 & 62.95\\
                IN21K $\to$ IN1K & MaxViT-B & 119M & 512x512 & 1024 & 63.42 & 51.07 & 49.87 & 44.10 & 12.53 & 77.67 & 54.35 & 60.69\\
                IN21K $\to$ IN1K & MaxViT-L & 212M & 512x512 & 1024 & \underline{67.27} & 52.70 & \underline{51.91} & \textbf{44.71} & \underline{13.09} & \underline{83.11} & 54.38 & \underline{63.84}\\
                IN21K $\to$ IN1K & MaxViT-XL & 475M & 512x512 & 1024 & \textbf{67.34} & 52.72 & 51.08 & 44.16 & 12.95 & \textbf{83.15} & 52.83 & 63.67\\
                \midrule
                \arrayrulecolor{black}
                IN21K & MambaVision-L2 & 242M & 512x512 & 1024 & 50.80 & 41.95 & 41.57 & 42.79 & 10.50 & 62.97 & 41.93 & 48.63\\
                IN21K & MambaVision-L3 & 740M & 256x256 & 256 & 52.18 & 43.58 & 36.27 & 42.94 & 10.66 & 65.13 & 38.51 & 49.26\\
                IN21K & MambaVision-L3 & 740M & 512x512 & 1024 & 51.50 & 42.62 & 40.91 & 43.40 & 10.45 & 61.98 & 43.19 & 49.32\\
                \midrule
                \arrayrulecolor{black}
                IN1K & VMamba-T & 30M & 224x224 & 196 & 64.99 & \underline{54.02} & 44.96 & 44.62 & 12.22 & 81.70 & 54.58 & 62.07\\
                IN1K & VMamba-S & 50M & 224x224 & 196 & 65.24 & \textbf{54.08} & 44.95 & 44.06 & 12.24 & 82.20 & \textbf{55.54} & 62.39\\
                IN1K & VMamba-B & 80M & 224x224 & 196 & 64.20 & 53.25 & 47.97 & 43.81 & 12.08 & 80.75 & 54.05 & 61.38\\
                \bottomrule
            \end{tabular}
            }
    \end{small}
  \end{center}
  \vskip -0.1in            
\end{table*}

\begin{table*}[!htbp]
    \caption{Additional localization results for selected MaxViT and MambaVision configurations beyond the matched 196-token setting. We report larger-resolution and larger-token variants for hierarchical backbones and compare them with the standard VMamba models.}

    \vskip -0.2in
    \label{tab:cls_unfair_res_pretrain-loc}
      \begin{center}
        \begin{small}
            \resizebox{\textwidth}{!}{%
            \begin{tabular}{c c c c c | c c c c | c || c} 
                \toprule
                Pretrained & Visual & Encoder & Image & Vision & \multirow{2}{*}{RefCOCO} & \multirow{2}{*}{RefCOCO+} & \multirow{2}{*}{RefCOCOg} & \multirow{2}{*}{OCID-Ref} & Weighted & Weighted \\
                Dataset & Encoder & Size & Size & Token \# &  &  &  &  & Loc. & Overall \\
                \midrule
                IN1K & MaxViT-T & 31M & 384x384 & 576 & 55.01 & 43.06 & 47.75 & 22.30 & 37.97 & 58.02\\
                IN1K & MaxViT-S & 69M & 384x384 & 576 & 25.68 & 14.97 & 18.22 & 4.08 & 13.46 & 52.62\\
                IN1K & MaxViT-B & 119M & 384x384 & 576 & 28.18 & 18.45 & 20.61 & 6.08 & 15.98 & 54.92\\
                IN1K & MaxViT-L & 212M & 384x384 & 576 & 23.92 & 13.65 & 17.12 & 5.36 & 13.12 & 53.46\\
                IN1K & MaxViT-T & 31M & 512x512 & 1024 & 54.54 & 43.28 & 48.24 & \textbf{23.42} & 38.42 & 58.03\\
                IN1K & MaxViT-S & 69M & 512x512 & 1024 & 35.07 & 23.51 & 28.10 & 9.29 & 20.99 & 55.37\\
                IN1K & MaxViT-B & 119M & 512x512 & 1024 & 33.23 & 22.73 & 25.14 & 8.34 & 19.64 & 55.92\\
                IN1K & MaxViT-L & 212M & 512x512 & 1024 & 25.60 & 15.34 & 18.83 & 5.90 & 14.34 & 53.72\\
                \cdashline{1-11}
                \arrayrulecolor{black}
                IN21K $\to$ IN1K & MaxViT-B & 119M & 384x384 & 576 & 21.22 & 12.01 & 14.56 & 2.87 & 10.77 & 54.00\\
                IN21K $\to$ IN1K & MaxViT-L & 212M & 384x384 & 576 & 26.46 & 15.92 & 19.36 & 5.79 & 14.70 & 57.28\\
                IN21K $\to$ IN1K & MaxViT-XL & 475M & 384x384 & 576 & 24.67 & 14.46 & 18.32 & 3.67 & 12.93 & 56.23\\
                IN21K $\to$ IN1K & MaxViT-B & 119M & 512x512 & 1024 & 21.72 & 12.49 & 15.26 & 2.58 & 10.97 & 54.01\\
                IN21K $\to$ IN1K & MaxViT-L & 212M & 512x512 & 1024 & 26.45 & 15.50 & 20.26 & 5.22 & 14.46 & 57.21\\
                IN21K $\to$ IN1K & MaxViT-XL & 475M & 512x512 & 1024 & 26.67 & 15.66 & 20.20 & 5.33 & 14.59 & 57.07\\
                \midrule
                \arrayrulecolor{black}
                IN21K & MambaVision-L2 & 242M & 512x512 & 1024 & 21.80 & 12.46 & 14.17 & 3.45 & 11.22 & 43.60\\
                IN21K & MambaVision-L3 & 740M & 256x256 & 256 & 21.39 & 11.89 & 13.93 & 3.71 & 11.06 & 44.13\\
                IN21K & MambaVision-L3 & 740M & 512x512 & 1024 & 21.27 & 11.61 & 13.46 & 3.66 & 10.89 & 44.15\\
                \midrule
                \arrayrulecolor{black}
                IN1K & VMamba-T & 30M & 224x224 & 196 & \textbf{58.25} & \textbf{46.64} & \textbf{51.74} & 20.24 & \textbf{39.20} & \underline{59.00}\\
                IN1K & VMamba-S & 50M & 224x224 & 196 & \underline{56.48} & \underline{44.27} & \underline{49.88} & \underline{23.09} & \underline{39.17} & \textbf{59.27}\\
                IN1K & VMamba-B & 80M & 224x224 & 196 & 42.06 & 31.43 & 36.15 & 15.23 & 27.89 & 56.88\\
                \bottomrule
            \end{tabular}
            }
    \end{small}
  \end{center}
  \vskip -0.2in            
\end{table*}

\newpage
\noindent\textit{Dense-objective pretraining beyond the controlled setup.}
Tables~\ref{tab:dense_unfair-vqa} and~\ref{tab:dense_unfair-loc} report additional segmentation-pretrained backbones. Under these much larger pretraining setups, very large models can achieve the strongest VQA results in this group. Nevertheless, VMamba variants remain consistently stronger on localization. These results reinforce the main paper's observation that dense objectives can improve downstream VLM performance, while also showing that better spatial transfer does not simply come from scaling model size, pretraining data, or resolution.\\

\begin{table*}[!htbp]
    \caption{VQA benchmarks for additional backbones with non-classification pretraining or task-adapted checkpoints, alongside VMamba transfer variants. Because these settings differ from the main controlled comparison in pretraining data and objective, they are reported separately.}

    \vskip -0.2in
    \label{tab:dense_unfair-vqa}
      \begin{center}
        \begin{small}
            \resizebox{\textwidth}{!}{%
            \begin{tabular}{c c c c c | c c c c c c c | c} 
                \toprule
                Pretrained & Visual & Encoder & Image & Vision & \multirow{2}{*}{VQA-v2} & \multirow{2}{*}{GQA} & \multirow{2}{*}{VizWiz} & TextVQA & \multirow{2}{*}{TextVQA} & \multirow{2}{*}{POPE} & \multirow{2}{*}{TallyQA} & Weighted \\
                Dataset & Encoder & Size & Size & Token \# &  &  &  & +OCR &  &  &  & VQA \\
                \midrule
                Multi-Modal $\to$ ADE20K & Perceiver & 364M & 512x512 & 256 & 65.62 & 54.63 & 46.25 & 44.26 & 12.25 & 83.79 & 56.62 & 62.92\\
                IN22K $\to$ ADE20K & BeiT-L & 451M & 640x640 & 400 & 68.78 & 56.52 & 48.06 & 44.75 & 13.01 & 86.88 & 58.24 & 65.70\\
                IN22K $\to$ ADE20K & BeiT-L & 568M & 640x640 & 400 & \underline{69.57} & 57.12 & \textbf{51.95} & 44.03 & \underline{13.14} & 86.57 & 58.64 & 66.41\\
                IN22K+COCO $\to$ ADE20K & BeiT-L & 571M & 896x896 & 784 & \underline{69.57} & \underline{57.47} & 49.73 & 44.58 & 12.89 & \underline{88.09} & 59.03 & \underline{66.49}\\
                IN22K+COCO $\to$ ADE20K & BeiTv2-L & 571M & 896x896 & 784 & \textbf{70.84} & \textbf{58.58} & 49.84 & 44.28 & \textbf{13.39} & \textbf{88.24} & \textbf{61.32} & \textbf{67.80}\\
                \cdashline{1-13}
                \arrayrulecolor{black}
                IN22K $\to$ ADE20K & AugReg-T & 36M & 512x512 & 256 & 58.52 & 50.25 & 47.98 & 43.92 & 11.53 & 76.08 & 52.32 & 56.65\\
                IN22K $\to$ ADE20K & AugReg-B & 134M & 512x512 & 256 & 61.51 & 52.68 & 46.41 & 44.65 & 11.99 & 78.28 & 54.20 & 59.29\\
                IN22K $\to$ ADE20K & AugReg-L & 364M & 512x512 & 256 & 65.92 & 55.60 & \underline{50.37} & 44.42 & 12.32 & 83.23 & 54.98 & 63.01\\
                \midrule
                \arrayrulecolor{black}
                IN1K $\to$ COCO & VMamba-T & 30M & 512x512 & 1024 & 66.91 & 55.30 & 45.05 & 44.43 & 11.73 & 84.86 & 58.94 & 64.22\\
                IN1K $\to$ COCO & VMamba-S & 50M & 512x512 & 1024 & 66.42 & 55.38 & 46.14 & \textbf{44.99} & 11.98 & 86.78 & 58.20 & 63.85\\
                IN1K $\to$ COCO & VMamba-B & 89M & 512x512 & 1024 & 65.70 & 55.01 & 46.07 & 44.66 & 12.05 & 87.99 & 60.07 & 63.58\\
                \cdashline{1-13}
                \arrayrulecolor{black}
                IN1K $\to$ COCO & VMamba-T(f) & 30M & 512x512 & 1024 & 66.77 & 55.41 & 42.75 & 44.28 & 11.72 & 85.71 & 60.19 & 64.28\\
                IN1K $\to$ COCO & VMamba-S(f) & 50M & 512x512 & 1024 & 65.98 & 55.18 & 45.69 & 44.44 & 12.02 & 87.28 & 60.42 & 63.81\\
                IN1K $\to$ COCO & VMamba-B(f) & 89M & 512x512 & 1024 & 65.63 & 55.11 & 44.13 & \underline{44.86} & 12.15 & 87.57 & \underline{60.69} & 63.58\\
                \cdashline{1-13}
                \arrayrulecolor{black}
                IN1K $\to$ ADE20K & VMamba-T & 30M & 512x512 & 1024 & 65.45 & 55.26 & 40.91 & 44.18 & 11.85 & 83.65 & 55.66 & 62.60\\
                IN1K $\to$ ADE20K & VMamba-S & 50M & 512x512 & 1024 & 66.42 & 55.68 & 47.44 & 44.42 & 11.86 & 84.01 & 53.91 & 63.21\\
                IN1K $\to$ ADE20K & VMamba-B & 89M & 512x512 & 1024 & 66.12 & 55.89 & 40.19 & 44.50 & 12.33 & 84.39 & 53.61 & 62.87\\
                \bottomrule
            \end{tabular}
            }
    \end{small}
  \end{center}
  \vskip -0.1in            
\end{table*}

\begin{table*}[!htbp] 
    \caption{Localization benchmarks for additional backbones with non-classification pretraining or task-adapted checkpoints, alongside VMamba transfer variants. Because these settings differ from the main controlled comparison in pretraining data and objective, they are reported separately.}

    \vskip -0.1in
    \label{tab:dense_unfair-loc}
      \begin{center}
        \begin{small}
            \resizebox{\textwidth}{!}{%
            \begin{tabular}{c c c c c | c c c c | c || c} 
                \toprule
                Pretrained & Visual & Encoder & Image & Vision & \multirow{2}{*}{RefCOCO} & \multirow{2}{*}{RefCOCO+} & \multirow{2}{*}{RefCOCOg} & \multirow{2}{*}{OCID-Ref} & Weighted & Weighted \\
                Dataset & Encoder & Size & Size & Token \# &  &  &  &  & Loc. & Overall \\
                \midrule
                Multi-Modal $\to$ ADE20K & Perceiver & 364M & 512x512 & 256 & 63.31 & 51.55 & 55.88 & 20.81 & 42.29 & 60.14\\
                IN22K $\to$ ADE20K & BeiT-L & 451M & 640x640 & 400 & 52.99 & 42.75 & 47.81 & 16.94 & 35.22 & 61.61\\
                IN22K $\to$ ADE20K & BeiT-L & 568M & 640x640 & 400 & 56.09 & 46.49 & 51.45 & 18.60 & 37.94 & 62.58\\
                IN22K+COCO $\to$ ADE20K & BeiT-L & 571M & 896x896 & 784 & 55.53 & 46.07 & 49.43 & 18.51 & 37.45 & \underline{62.59}\\
                IN22K+COCO $\to$ ADE20K & BeiTv2-L & 571M & 896x896 & 784 & 60.34 & 49.29 & 53.33 & 24.69 & 42.34 & \textbf{64.38}\\
                \cdashline{1-11}
                \arrayrulecolor{black}
                IN22K $\to$ ADE20K & AugReg-T & 36M & 512x512 & 256 & 41.19 & 29.14 & 34.17 & 11.32 & 25.31 & 52.44\\
                IN22K $\to$ ADE20K & AugReg-B & 134M & 512x512 & 256 & 48.32 & 36.36 & 41.44 & 15.54 & 31.29 & 55.53\\
                IN22K $\to$ ADE20K & AugReg-L & 364M & 512x512 & 256 & 56.58 & 45.16 & 50.47 & 20.28 & 38.32 & 59.69\\
                \midrule
                \arrayrulecolor{black}
                IN1K $\to$ COCO & VMamba-T & 30M & 512x512 & 1024 & 61.51 & 50.50 & 53.76 & 28.50 & 44.52 & 61.57\\
                IN1K $\to$ COCO & VMamba-S & 50M & 512x512 & 1024 & \underline{64.64} & 52.22 & 56.35 & \textbf{32.50} & \textbf{47.60} & 61.67\\
                IN1K $\to$ COCO & VMamba-B & 89M & 512x512 & 1024 & 62.77 & 50.14 & 56.33 & 30.00 & 45.63 & 61.17\\
                \cdashline{1-11}
                \arrayrulecolor{black}
                IN1K $\to$ COCO & VMamba-T(f) & 30M & 512x512 & 1024 & 63.69 & 52.70 & 55.68 & 30.87 & 46.75 & 61.92\\
                IN1K $\to$ COCO & VMamba-S(f) & 50M & 512x512 & 1024 & \textbf{65.09} & \underline{52.81} & 57.13 & 31.14 & \underline{47.38} & 61.60\\
                IN1K $\to$ COCO & VMamba-B(f) & 89M & 512x512 & 1024 & 64.57 & 50.99 & 57.07 & \underline{31.35} & 46.90 & 61.34\\
                \cdashline{1-11}
                \arrayrulecolor{black}
                IN1K $\to$ ADE20K & VMamba-T & 30M & 512x512 & 1024 & 62.65 & 51.10 & 57.15 & 24.01 & 43.47 & 60.03\\
                IN1K $\to$ ADE20K & VMamba-S & 50M & 512x512 & 1024 & 64.17 & \textbf{53.98} & \textbf{59.13} & 24.58 & 44.98 & 60.76\\
                IN1K $\to$ ADE20K & VMamba-B & 89M & 512x512 & 1024 & 63.99 & 52.52 & \underline{58.68} & 25.91 & 45.08 & 60.48\\
                \bottomrule
            \end{tabular}
            }
    \end{small}
  \end{center}
  \vskip -0.1in            
\end{table*}

\newpage

\begin{table*}[!htbp]
    \caption{VQA benchmarks for backbones pretrained with contrastive or self-supervised objectives, along with fused DINOv2/CLIP and DINOv2/SigLIP variants. Rows marked (f) use a stronger connector for stabilization.}
    \vskip -0.1in
    \label{tab:obj_unfair-vqa}
      \begin{center}
        \begin{small}
            \resizebox{\textwidth}{!}{%
            \begin{tabular}{c c c c | c c c c c c c | c} 
                \toprule
                Visual & Encoder & Image & Vision & \multirow{2}{*}{VQA-v2} & \multirow{2}{*}{GQA} & \multirow{2}{*}{VizWiz} & TextVQA & \multirow{2}{*}{TextVQA} & \multirow{2}{*}{POPE} & \multirow{2}{*}{TallyQA} & Weighted \\
                Encoder & Size & Size & Token \# &  &  &  & +OCR &  &  &  & VQA \\
                \midrule
                CLIP-B & 86M & 224x224 & 196 & 72.25 & 58.41 & 52.92 & 51.01 & 31.98 & 85.04 & 60.65 & 69.14\\
                CLIP-L & 304M & 224x224 & 256 & 74.69 & 59.15 & 53.73 & 52.87 & 37.29 & 86.20 & 60.49 & 71.13\\
                DINOv2-L & 304M & 224x224 & 256 & 64.15 & 52.37 & 46.30 & 44.37 & 12.67 & 82.18 & 56.58 & 61.68\\
                SigLIP-B16 & 93M & 224x224 & 196 & 72.52 & 58.26 & 44.73 & 50.70 & 31.48 & 83.62 & 61.16 & 69.22\\
                SigLIP-SO400M & 428M & 224x224 & 256 & 75.88 & 59.76 & 52.79 & 55.09 & 42.13 & 84.93 & 64.50 & 72.65\\
                SigLIP-B16 & 93M & 256x256 & 256 & 73.30 & 58.67 & 46.70 & 52.39 & 34.98 & 83.84 & 62.11 & 70.07\\
                CLIP-L & 304M & 336x336 & 576 & 76.63 & 60.19 & 52.98 & 57.05 & 45.78 & 87.55 & 64.43 & 73.39\\
                SigLIP-B16 & 93M & 384x384 & 576 & 75.77 & 60.05 & 47.55 & 55.64 & 44.37 & 86.13 & 62.88 & 72.37\\
                SigLIP-SO400M & 428M & 384x384 & 729 & \underline{78.42} & \underline{61.09} & 56.04 & \textbf{60.79} & \textbf{55.46} & 87.69 & \underline{67.80} & \underline{75.49}\\
                \midrule
                \arrayrulecolor{black}
                DINOv2-L/SigLIP-SO400M & 304M/428M & 224x224/224x224 & 256+256 & 69.03 & 54.49 & 52.62 & 49.64 & 27.30 & 83.97 & 60.79 & 66.45 (\textcolor{red}{-6.58})\\
                DINOv2-L/CLIP-L & 304M/304M & 336x336/336x336 & 576+576 & 65.08 & 52.50 & 47.65 & 45.29 & 12.93 & 81.98 & 58.03 & 62.60 (\textcolor{red}{-8.34})\\
                DINOv2-L/SigLIP-SO400M & 304M/428M & 384x384/384x384 & 729+729 & 67.83 & 52.95 & 51.29 & 50.15 & 28.31 & 83.39 & 60.30 & 65.42 (\textcolor{red}{-10.33})\\
                \cdashline{1-12}
                \arrayrulecolor{black}
                \rowcolor{blue!8}
                DINOv2-L/SigLIP-SO400M(f) & 304M/428M & 224x224/224x224 & 256+256 & 76.09 & 60.24 & \underline{56.29} & 54.48 & 39.41 & 86.47 & 65.75 & 73.03\\
                \rowcolor{blue!8}
                DINOv2-L/CLIP-L(f) & 304M/304M & 336x336/336x336 & 576+576 & 74.34 & 60.96 & 52.69 & 45.96 & 15.22 & \underline{88.09} & 63.84 & 70.94\\
                \rowcolor{blue!8}
                DINOv2-L/SigLIP-SO400M(f) & 304M/428M & 384x384/384x384 & 729+729 & \textbf{78.59} & \textbf{61.66} & \textbf{59.36} & \underline{59.93} & \underline{52.58} & \textbf{88.31} & \textbf{68.52} & \textbf{75.75}\\
                \bottomrule
            \end{tabular}
            }
    \end{small}
  \end{center}
  \vskip -0.1in            
\end{table*}

\begin{table*}[!htbp]
    \caption{Localization benchmarks for backbones pretrained with contrastive or self-supervised objectives, along with fused DINOv2/CLIP and DINOv2/SigLIP variants. Rows marked (f) use a stronger connector for stabilization. Direct fusion can hurt localization, while stabilized fusion substantially improves grounding performance and yields the strongest results in this group.}
    \vskip -0.1in
    \label{tab:obj_unfair-loc}
      \begin{center}
        \begin{small}
            \resizebox{\textwidth}{!}{%
            \begin{tabular}{c c c c | c c c c | c || c} 
                \toprule
                Visual & Encoder & Image & Vision & \multirow{2}{*}{RefCOCO} & \multirow{2}{*}{RefCOCO+} & \multirow{2}{*}{RefCOCOg} & \multirow{2}{*}{OCID-Ref} & Weighted & Weighted \\
                Encoder & Size & Size & Token \# &  &  &  &  & Loc. & Overall \\
                \midrule
                CLIP-B & 86M & 224x224 & 196 & 64.85 & 55.57 & 60.05 & 28.40 & 47.19 & 66.19\\
                CLIP-L & 304M & 224x224 & 256 & 68.05 & 58.26 & 62.79 & 32.70 & 50.66 & 68.38\\
                DINOv2-L & 304M & 224x224 & 256 & 27.53 & 16.46 & 18.40 & 4.07 & 14.28 & 55.31\\
                SigLIP-B16 & 93M & 224x224 & 196 & 63.85 & 54.60 & 60.36 & 23.78 & 44.85 & 65.94\\
                SigLIP-SO400M & 428M & 224x224 & 256 & 63.26 & 54.39 & 57.60 & 23.61 & 44.29 & 68.84\\
                SigLIP-B16 & 93M & 256x256 & 256 & 64.77 & 55.78 & 60.09 & 25.62 & 46.08 & 66.85\\
                CLIP-L & 304M & 336x336 & 576 & 72.31 & 63.81 & 68.69 & 36.19 & 55.10 & 70.93\\
                SigLIP-B16 & 93M & 384x384 & 576 & 71.57 & 62.79 & 66.99 & 34.81 & 53.92 & 69.89\\
                SigLIP-SO400M & 428M & 384x384 & 729 & 69.17 & 60.98 & 66.89 & 34.46 & 52.75 & \underline{72.44}\\
                \midrule
                \arrayrulecolor{black}
                DINOv2-L/SigLIP-SO400M & 304M/428M & 224x224/224x224 & 256+256 & 33.07 & 22.33 & 26.51 & 4.04 & 17.90 (\textcolor{red}{-35.66}) & 59.93 (\textcolor{red}{-10.49})\\
                DINOv2-L/CLIP-L & 304M/304M & 336x336/336x336 & 576+576 & 27.99 & 15.85 & 18.85 & 3.16 & 13.92 (\textcolor{red}{-45.48}) & 56.06 (\textcolor{red}{-13.33})\\
                DINOv2-L/SigLIP-SO400M & 304M/428M & 384x384/384x384 & 729+729 & 29.45 & 17.24 & 21.81 & 4.29 & 15.39 (\textcolor{red}{-48.99}) & 58.69 (\textcolor{red}{-15.53})\\
                \cdashline{1-10}
                \arrayrulecolor{black}
                \rowcolor{blue!8}
                DINOv2-L/SigLIP-SO400M(f) & 304M/428M & 224x224/224x224 & 256+256 & 69.87 & 61.21 & 65.83 & 36.17 & 53.56 & 70.42\\
                \rowcolor{blue!8}
                DINOv2-L/CLIP-L(f) & 304M/304M & 336x336/336x336 & 576+576 & \underline{75.94} & \underline{66.57} & \underline{72.18} & \underline{42.01} & \underline{59.40} & 69.39\\
                \rowcolor{blue!8}
                DINOv2-L/SigLIP-SO400M(f) & 304M/428M & 384x384/384x384 & 729+729 & \textbf{78.83} & \textbf{70.25} & \textbf{75.29} & \textbf{49.50} & \textbf{64.38} & \textbf{74.22}\\
                \bottomrule
            \end{tabular}
            }
    \end{small}
  \end{center}
  \vskip -0.1in            
\end{table*}

\subsection{Contrastive/Self-Supervised Backbones}
\noindent\textit{Alternative pretraining objectives.}
Tables~\ref{tab:obj_unfair-vqa} and~\ref{tab:obj_unfair-loc} report additional backbones pretrained with contrastive or self-supervised objectives, as well as fused encoders built from them. Large contrastive and self-supervised backbones already achieve strong performance in this setting, especially when paired with larger models and higher input resolutions. Direct fusion, however, is not always stable, particularly for localization. Once stabilized with a stronger connector, the fused DINOv2 and SigLIP models become the strongest results in this group, supporting our observation that combining complementary pretraining objectives can further improve VLM performance. At the same time, these models use substantially larger vision stacks and higher token budgets than the standard VMamba setting, so we report them separately from the main controlled comparison. Overall, these results suggest that bringing contrastive or self-supervised objectives to VMamba-style backbones is a promising direction for future work.

\section{Single-GPU Inference Profiling}

\subsection{Profiling Setup}
We profile inference efficiency for representative ViT, VMamba, and ViTDet backbones inside the same VLM wrapper on a single GPU. We choose these three families to cover different practical trade-offs. ViT provides a standard transformer baseline at a scale similar to VMamba, which allows a direct comparison between two backbone families with similar model size but different VLM performance. ViTDet is included as a larger and competitive alternative, allowing us to compare against a higher-capacity vision backbone with a different runtime and memory profile.

For each backbone and input resolution, we rebuild the vision encoder at the target square resolution, combine it with the same projector and LLM, and run a full multimodal forward pass under \texttt{torch.inference\_mode()}. To focus on how inference cost changes with resolution, we use synthetic inputs instead of dataset samples. Unless otherwise stated, the batch size is 1 and the text length is fixed to 128 tokens. Each configuration is run for 200 forward iterations. We treat the first 100 iterations as warm-up and report the average of the final 100. For each model, we sweep resolutions in ascending order and stop at the first out-of-memory setting.

\begin{figure}[h!]
    \centering
    \includegraphics[width=1\linewidth]{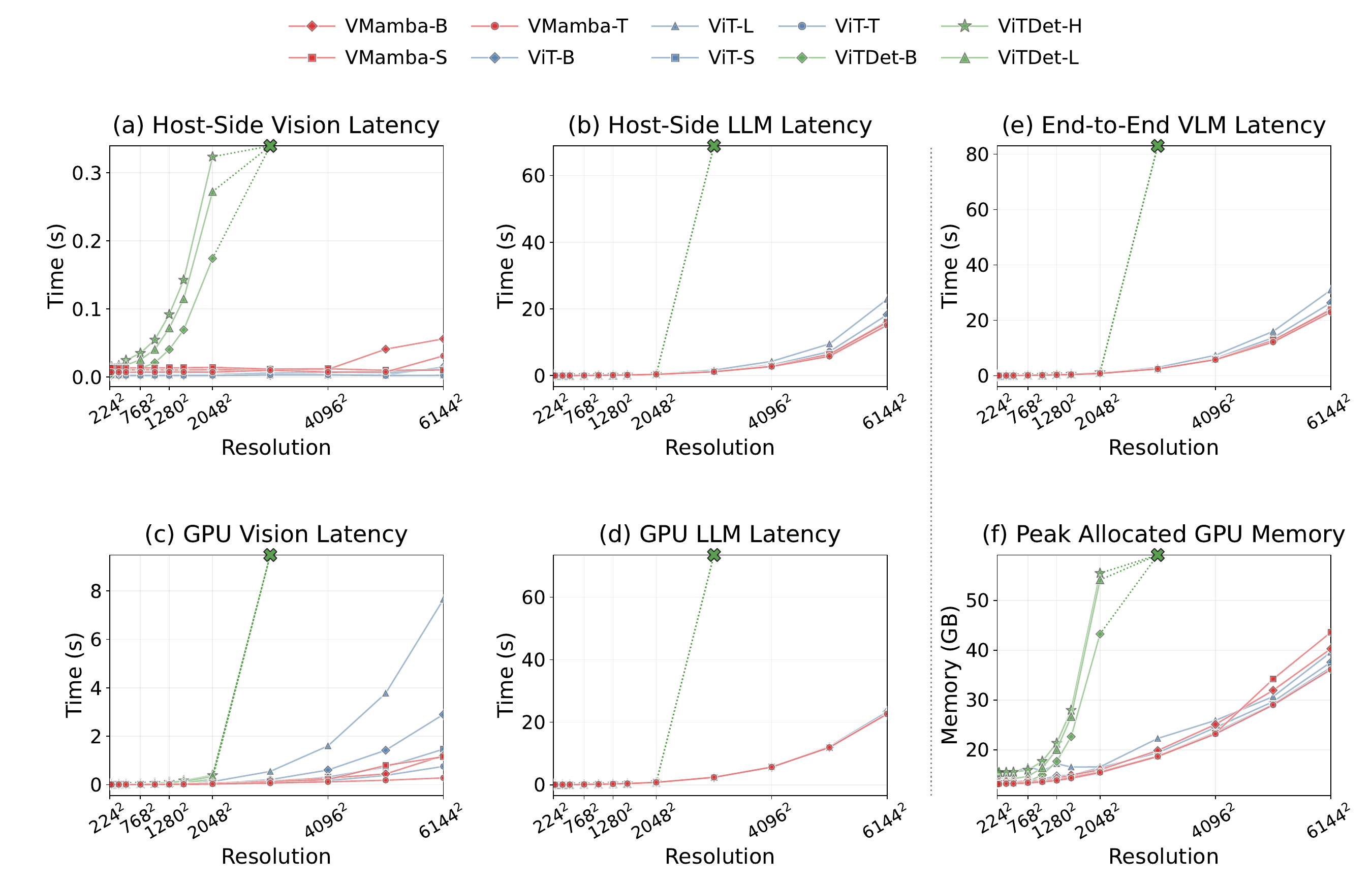}
    \vspace{-0.7cm}
    \caption{\textbf{Scaling of single-GPU VLM inference cost with image resolution at batch size 1.} The x-axis shows the square input resolution, written as $R^2$. The six panels report (a) host-side vision latency, (b) host-side LLM latency, (c) GPU vision latency, (d) GPU LLM latency, (e) end-to-end VLM latency, and (f) peak allocated GPU memory. For each model and resolution, we run 100 warm-up iterations and report the mean over the next 100 iterations. Dotted segments ending with an $\times$ indicate the first resolution that exceeds GPU memory capacity. The marker is placed at the top of each subplot to show the out-of-memory threshold rather than a measured metric value. In panel (f), it therefore marks the first failed resolution, not an observed peak-memory measurement.}
    \label{fig:profile}
\end{figure}

We report six metrics in Fig.~\ref{fig:profile}. \textbf{Host-side vision latency} measures the wall-clock time of the vision stage, including the vision backbone and projector. \textbf{Host-side LLM latency} measures the wall-clock time of the language stage, including multimodal input preparation and the LLM forward pass. \textbf{GPU vision latency} and \textbf{GPU LLM latency} measure how much time the GPU spends on the vision and language stages, respectively. \textbf{End-to-end VLM latency} measures the total wall-clock time of one multimodal forward pass. \textbf{Peak allocated GPU memory} records the largest GPU memory allocation during the forward pass.

These metrics capture different views of inference cost. The host-side timings reflect the overall time seen from the CPU side, so they include both model execution and the overhead of preparing and launching the computation. The GPU timings focus on the computation time spent on the GPU. Because these two views measure different parts of the system, they do not need to match exactly, and their sums do not necessarily equal the end-to-end latency.

In practice, end-to-end VLM latency is the most direct measure of inference speed. The stagewise GPU timings show where the main computation cost lies, the host-side timings show the overall runtime overhead of each stage, and peak memory determines the largest resolution that can be run. Ignoring the first 100 iterations reduces startup effects and gives a more stable estimate of steady-state inference cost.

\subsection{Observations}
Figure~\ref{fig:profile} shows a clear practical trade-off across the three backbone families. ViT and VMamba operate in a broadly similar efficiency regime, while ViTDet is substantially heavier and reaches the memory limit much earlier. This makes the comparison informative: ViT serves as a scale-matched baseline for VMamba, and ViTDet shows the cost of using a much larger vision backbone.

First, Fig.~\ref{fig:profile}(f) shows that ViTDet runs out of memory at much lower resolutions than ViT and VMamba. This is consistent with its much larger model size and indicates that memory, rather than latency alone, quickly becomes the limiting factor for this family.\\

Second, Fig.~\ref{fig:profile}(c) shows that GPU vision latency grows more rapidly for ViT as the input resolution increases, whereas VMamba scales more gradually. This suggests that VMamba handles higher-resolution visual inputs more efficiently in the vision stage. Fig.~\ref{fig:profile}(a) shows a similar trend on the host side: vision-stage overhead remains relatively small for ViT and VMamba, but becomes much larger for the heavier ViTDet models.

Third, Fig.~\ref{fig:profile}(b), (d), and (e) show that the language stage dominates the overall inference time for most settings. For ViT and VMamba, the end-to-end latency curves are much closer to the LLM-stage curves than to the vision-stage curves, which indicates that the vision module is often a smaller part of total VLM inference cost at this scale. The vision-stage differences become more visible mainly at very high resolutions or for much larger backbones such as ViTDet.

Overall, ViT and VMamba have similar practical inference cost at comparable scale, but VMamba achieves stronger VLM performance in our main results. ViTDet is also competitive in VLM performance, but it is much larger and reaches out-of-memory much earlier. Taken together, these results make VMamba an attractive design point: it offers a stronger performance--efficiency trade-off than a similarly sized ViT baseline, while leaving more room than ViTDet for future scaling and system design choices.

\vspace{-3pt}
\section{Exhaustive Results}
\vspace{-3pt}
\label{app:exhaustive_results}

For completeness, we provide exhaustive VQA and localization results for all evaluated backbone families. Specifically, Tables~\ref{tab:exhaustive-VMamba-vqa} and~\ref{tab:exhaustive-VMamba-loc} report VMamba, Tables~\ref{tab:exhaustive-ViT-vqa} and~\ref{tab:exhaustive-ViT-loc} report ViT, Tables~\ref{tab:exhaustive-MaxViT-vqa} and~\ref{tab:exhaustive-MaxViT-loc} report MaxViT, Tables~\ref{tab:exhaustive-MambaVision-vqa} and~\ref{tab:exhaustive-MambaVision-loc} report MambaVision, Tables~\ref{tab:exhaustive-Vim-vqa} and~\ref{tab:exhaustive-Vim-loc} report Vim, Tables~\ref{tab:exhaustive-ViTDet-vqa} and~\ref{tab:exhaustive-ViTDet-loc} report ViTDet, and Tables~\ref{tab:exhaustive-ViT-Adapter-vqa} and~\ref{tab:exhaustive-ViT-Adapter-loc} report ViT-Adapter.

\vspace{-4pt}
\begin{table*}[!htbp]
    \caption{VMamba VQA benchmarks}
    \vskip -0.1in
    \label{tab:exhaustive-VMamba-vqa}
      \begin{center}
        \begin{small}
            \resizebox{\textwidth}{!}{%
            \begin{tabular}{c c c c c | c c c c c c c | c} 
                \toprule
                Pretrained & Visual & Encoder & Image & Vision & \multirow{2}{*}{VQA-v2} & \multirow{2}{*}{GQA} & \multirow{2}{*}{VizWiz} & TextVQA & \multirow{2}{*}{TextVQA} & \multirow{2}{*}{POPE} & \multirow{2}{*}{TallyQA} & Weighted \\
                Dataset & Encoder & Size & Size & Token \# &  &  &  & +OCR &  &  &  & VQA \\
                \midrule
                IN1K & VMamba-T & 30M & 224x224 & 196 & 64.99 & 54.02 & 44.96 & 44.62 & 12.22 & 81.70 & 54.58 & 62.07\\
                IN1K & VMamba-S & 50M & 224x224 & 196 & 65.24 & 54.08 & 44.95 & 44.06 & 12.24 & 82.20 & 55.54 & 62.39\\
                IN1K & VMamba-B & 80M & 224x224 & 196 & 64.20 & 53.25 & 47.97 & 43.81 & 12.08 & 80.75 & 54.05 & 61.38\\
                \midrule
                \arrayrulecolor{black}
                IN1K & VMamba-T(f) & 30M & 224x224 & 196 & 65.22 & 53.87 & 44.88 & 44.84 & 12.19 & 81.19 & 56.29 & 62.45\\
                IN1K & VMamba-S(f) & 50M & 224x224 & 196 & 65.49 & 54.75 & 47.40 & 43.77 & \textbf{12.40} & 81.70 & 55.93 & 62.68\\
                IN1K & VMamba-B(f) & 80M & 224x224 & 196 & 64.51 & 53.81 & \underline{48.83} & 43.97 & \underline{12.36} & 81.83 & 56.38 & 62.00\\
                \midrule
                \arrayrulecolor{black}
                IN1K $\to$ COCO & VMamba-T & 30M & 1333x800 & 4150 & 60.24 & 49.18 & \textbf{49.04} & 44.06 & 11.81 & 79.48 & 52.65 & 58.05\\
                IN1K $\to$ COCO & VMamba-S & 50M & 1333x800 & 4150 & 64.87 & 53.98 & 47.73 & 44.24 & 11.93 & 86.47 & 59.21 & 62.78\\
                IN1K $\to$ COCO & VMamba-B & 89M & 1333x800 & 4150 & 60.36 & 49.21 & 46.46 & 43.84 & 11.73 & 81.61 & 55.69 & 58.57\\
                \midrule
                \arrayrulecolor{black}
                IN1K $\to$ COCO & VMamba-T(f) & 30M & 1333x800 & 4150 & 64.46 & 53.33 & 45.21 & 43.68 & 11.90 & 84.14 & 54.27 & 61.66\\
                IN1K $\to$ COCO & VMamba-S(f) & 50M & 1333x800 & 4150 & 64.31 & 53.84 & 43.31 & 44.20 & 12.23 & 86.36 & 57.13 & 62.01\\
                IN1K $\to$ COCO & VMamba-B(f) & 89M & 1333x800 & 4150 & 60.85 & 50.25 & 40.44 & 43.99 & 11.77 & 82.71 & 55.04 & 58.84\\
                \midrule
                \arrayrulecolor{black}
                IN1K $\to$ COCO & VMamba-T & 30M & 512x512 & 1024 & \textbf{66.91} & 55.30 & 45.05 & 44.43 & 11.73 & 84.86 & 58.94 & \underline{64.22}\\
                IN1K $\to$ COCO & VMamba-S & 50M & 512x512 & 1024 & 66.42 & 55.38 & 46.14 & \textbf{44.99} & 11.98 & 86.78 & 58.20 & 63.85\\
                IN1K $\to$ COCO & VMamba-B & 89M & 512x512 & 1024 & 65.70 & 55.01 & 46.07 & 44.66 & 12.05 & \textbf{87.99} & 60.07 & 63.58\\
                \midrule
                \arrayrulecolor{black}
                IN1K $\to$ COCO & VMamba-T(f) & 30M & 512x512 & 1024 & \underline{66.77} & 55.41 & 42.75 & 44.28 & 11.72 & 85.71 & 60.19 & \textbf{64.28}\\
                IN1K $\to$ COCO & VMamba-S(f) & 50M & 512x512 & 1024 & 65.98 & 55.18 & 45.69 & 44.44 & 12.02 & 87.28 & \underline{60.42} & 63.81\\
                IN1K $\to$ COCO & VMamba-B(f) & 89M & 512x512 & 1024 & 65.63 & 55.11 & 44.13 & \underline{44.86} & 12.15 & \underline{87.57} & \textbf{60.69} & 63.58\\
                \midrule
                \arrayrulecolor{black}
                IN1K $\to$ ADE20K & VMamba-T & 30M & 512x512 & 1024 & 65.45 & 55.26 & 40.91 & 44.18 & 11.85 & 83.65 & 55.66 & 62.60\\
                IN1K $\to$ ADE20K & VMamba-S & 50M & 512x512 & 1024 & 66.42 & \underline{55.68} & 47.44 & 44.42 & 11.86 & 84.01 & 53.91 & 63.21\\
                IN1K $\to$ ADE20K & VMamba-B & 89M & 512x512 & 1024 & 66.12 & \textbf{55.89} & 40.19 & 44.50 & 12.33 & 84.39 & 53.61 & 62.87\\
                \bottomrule
            \end{tabular}
            }
    \end{small}
  \end{center}
  \vskip -0.1in            
\end{table*}

\begin{table*}[!htbp]
    \caption{VMamba localization benchmarks}
    \vskip -0.1in
    \label{tab:exhaustive-VMamba-loc}
      \begin{center}
        \begin{small}
            \resizebox{\textwidth}{!}{%
            \begin{tabular}{c c c c c | c c c c | c || c} 
                \toprule
                Pretrained & Visual & Encoder & Image & Vision & \multirow{2}{*}{RefCOCO} & \multirow{2}{*}{RefCOCO+} & \multirow{2}{*}{RefCOCOg} & \multirow{2}{*}{OCID-Ref} & Weighted & Weighted \\
                Dataset & Encoder & Size & Size & Token \# &  &  &  &  & Loc. & Overall \\
                \midrule
                IN1K & VMamba-T & 30M & 224x224 & 196 & 58.25 & 46.64 & 51.74 & 20.24 & 39.20 & 59.00\\
                IN1K & VMamba-S & 50M & 224x224 & 196 & 56.48 & 44.27 & 49.88 & 23.09 & 39.17 & 59.27\\
                IN1K & VMamba-B & 80M & 224x224 & 196 & 42.06 & 31.43 & 36.15 & 15.23 & 27.89 & 56.88\\
                \midrule
                \arrayrulecolor{black}
                IN1K & VMamba-T(f) & 30M & 224x224 & 196 & 58.22 & 47.30 & 51.76 & 22.00 & 40.07 & 59.44\\
                IN1K & VMamba-S(f) & 50M & 224x224 & 196 & 54.65 & 43.94 & 49.24 & 24.35 & 39.09 & 59.51\\
                IN1K & VMamba-B(f) & 80M & 224x224 & 196 & 45.35 & 33.96 & 39.32 & 16.82 & 30.29 & 57.74\\
                \midrule
                \arrayrulecolor{black}
                IN1K $\to$ COCO & VMamba-T & 30M & 1333x800 & 4150 & 29.10 & 16.83 & 19.89 & 3.95 & 14.86 & 52.25\\
                IN1K $\to$ COCO & VMamba-S & 50M & 1333x800 & 4150 & \textbf{69.52} & \underline{52.87} & \textbf{63.50} & 28.15 & \textbf{47.94} & 60.78\\
                IN1K $\to$ COCO & VMamba-B & 89M & 1333x800 & 4150 & 28.43 & 16.44 & 19.87 & 4.97 & 15.02 & 52.72\\
                \midrule
                \arrayrulecolor{black}
                IN1K $\to$ COCO & VMamba-T(f) & 30M & 1333x800 & 4150 & 57.10 & 42.55 & 50.10 & 20.64 & 37.93 & 58.47\\
                IN1K $\to$ COCO & VMamba-S(f) & 50M & 1333x800 & 4150 & \underline{67.40} & 51.02 & \underline{61.38} & 28.89 & 47.06 & 60.00\\
                IN1K $\to$ COCO & VMamba-B(f) & 89M & 1333x800 & 4150 & 31.47 & 19.10 & 22.24 & 3.95 & 16.23 & 53.11\\
                \midrule
                \arrayrulecolor{black}
                IN1K $\to$ COCO & VMamba-T & 30M & 512x512 & 1024 & 61.51 & 50.50 & 53.76 & 28.50 & 44.52 & 61.57\\
                IN1K $\to$ COCO & VMamba-S & 50M & 512x512 & 1024 & 64.64 & 52.22 & 56.35 & \textbf{32.50} & \underline{47.60} & \underline{61.67}\\
                IN1K $\to$ COCO & VMamba-B & 89M & 512x512 & 1024 & 62.77 & 50.14 & 56.33 & 30.00 & 45.63 & 61.17\\
                \midrule
                \arrayrulecolor{black}
                IN1K $\to$ COCO & VMamba-T(f) & 30M & 512x512 & 1024 & 63.69 & 52.70 & 55.68 & 30.87 & 46.75 & \textbf{61.92}\\
                IN1K $\to$ COCO & VMamba-S(f) & 50M & 512x512 & 1024 & 65.09 & 52.81 & 57.13 & 31.14 & 47.38 & 61.60\\
                IN1K $\to$ COCO & VMamba-B(f) & 89M & 512x512 & 1024 & 64.57 & 50.99 & 57.07 & \underline{31.35} & 46.90 & 61.34\\
                \midrule
                \arrayrulecolor{black}
                IN1K $\to$ ADE20K & VMamba-T & 30M & 512x512 & 1024 & 62.65 & 51.10 & 57.15 & 24.01 & 43.47 & 60.03\\
                IN1K $\to$ ADE20K & VMamba-S & 50M & 512x512 & 1024 & 64.17 & \textbf{53.98} & 59.13 & 24.58 & 44.98 & 60.76\\
                IN1K $\to$ ADE20K & VMamba-B & 89M & 512x512 & 1024 & 63.99 & 52.52 & 58.68 & 25.91 & 45.08 & 60.48\\
                \bottomrule
            \end{tabular}
            }
    \end{small}
  \end{center}
  \vskip -0.1in            
\end{table*}

\begin{table*}[t] 
    \caption{ViT VQA benchmarks}
    \vskip -0.1in
    \label{tab:exhaustive-ViT-vqa}
      \begin{center}
        \begin{small}
            \resizebox{\textwidth}{!}{%
            \begin{tabular}{c c c c c | c c c c c c c | c} 
                \toprule
                Pretrained & Visual & Encoder & Image & Vision & \multirow{2}{*}{VQA-v2} & \multirow{2}{*}{GQA} & \multirow{2}{*}{VizWiz} & TextVQA & \multirow{2}{*}{TextVQA} & \multirow{2}{*}{POPE} & \multirow{2}{*}{TallyQA} & Weighted \\
                Dataset & Encoder & Size & Size & Token \# &  &  &  & +OCR &  &  &  & VQA \\
                \midrule
                IN21k & ViT-T & 5.7M & 224x224 & 196 & 62.82 & 51.30 & 46.03 & \underline{44.62} & 11.90 & 76.94 & 52.65 & 59.95\\
                IN21k & ViT-S & 22M & 224x224 & 196 & 61.28 & 50.34 & 47.95 & 44.18 & 12.30 & 78.59 & 53.62 & 58.97\\
                IN21k & ViT-B & 87M & 224x224 & 196 & \textbf{68.16} & \textbf{55.96} & \textbf{49.40} & \textbf{44.75} & \textbf{12.77} & \underline{82.34} & \underline{55.26} & \textbf{64.70}\\
                IN21k & ViT-L & 304M & 224x224 & 196 & 67.13 & 54.85 & \underline{49.15} & 44.28 & 12.64 & \textbf{82.38} & 53.32 & 63.61\\
                \midrule
                \arrayrulecolor{black}
                IN21k $\to$ IN1K & ViT-T & 5.7M & 224x224 & 196 & 61.34 & 50.00 & 45.30 & 44.02 & 11.84 & 77.90 & 53.82 & 58.96\\
                IN21k $\to$ IN1K & ViT-S & 22M & 224x224 & 196 & 60.39 & 49.46 & 44.54 & 44.58 & 12.23 & 78.49 & 51.60 & 57.95\\
                IN21k $\to$ IN1K & ViT-B & 87M & 224x224 & 196 & \underline{67.24} & \underline{54.96} & 47.76 & 43.92 & 12.34 & 82.16 & \textbf{57.16} & \underline{64.17}\\
                IN21k $\to$ IN1K & ViT-L & 304M & 224x224 & 196 & 62.17 & 50.82 & 42.10 & 44.40 & \underline{12.71} & 78.42 & 53.47 & 59.55\\
                \midrule
                \arrayrulecolor{black}
                IN1K & ViT-S & 22M & 224x224 & 196 & 59.86 & 50.48 & 46.69 & 44.02 & 12.30 & 77.76 & 48.96 & 57.25\\
                IN1K & ViT-B & 87M & 224x224 & 196 & 59.63 & 49.19 & 45.63 & 44.51 & 12.07 & 75.90 & 50.57 & 57.17\\
                \bottomrule
            \end{tabular}
            }
    \end{small}
  \end{center}
  \vskip -0.1in            
\end{table*}

\begin{table*}[t] 
    \caption{ViT localization benchmarks}
    \vskip -0.1in
    \label{tab:exhaustive-ViT-loc}
      \begin{center}
        \begin{small}
            \resizebox{\textwidth}{!}{%
            \begin{tabular}{c c c c c | c c c c | c || c} 
                \toprule
                Pretrained & Visual & Encoder & Image & Vision & \multirow{2}{*}{RefCOCO} & \multirow{2}{*}{RefCOCO+} & \multirow{2}{*}{RefCOCOg} & \multirow{2}{*}{OCID-Ref} & Weighted & Weighted \\
                Dataset & Encoder & Size & Size & Token \# &  &  &  &  & Loc. & Overall \\
                \midrule
                IN21k & ViT-T & 5.7M & 224x224 & 196 & 34.52 & 21.72 & 26.47 & 7.30 & 19.43 & 54.50\\
                IN21k & ViT-S & 22M & 224x224 & 196 & 24.40 & 13.79 & 17.42 & 4.73 & 13.04 & 52.80\\
                IN21k & ViT-B & 87M & 224x224 & 196 & \textbf{55.00} & \underline{44.79} & \textbf{48.35} & \textbf{19.90} & \textbf{37.46} & \textbf{61.04}\\
                IN21k & ViT-L & 304M & 224x224 & 196 & 40.64 & 31.38 & 35.21 & 11.40 & 25.86 & 58.54\\
                \midrule
                \arrayrulecolor{black}
                IN21k $\to$ IN1K & ViT-T & 5.7M & 224x224 & 196 & 31.56 & 19.84 & 23.79 & 8.35 & 18.40 & 53.51\\
                IN21k $\to$ IN1K & ViT-S & 22M & 224x224 & 196 & 24.29 & 14.05 & 16.99 & 4.05 & 12.75 & 51.88\\
                IN21k $\to$ IN1K & ViT-B & 87M & 224x224 & 196 & \underline{54.82} & \textbf{45.00} & \underline{48.30} & \underline{18.44} & \underline{36.87} & \underline{60.50}\\
                IN21k $\to$ IN1K & ViT-L & 304M & 224x224 & 196 & 22.92 & 13.04 & 15.77 & 3.18 & 11.69 & 53.12\\
                \midrule
                \arrayrulecolor{black}
                IN1K & ViT-S & 22M & 224x224 & 196 & 32.32 & 21.77 & 24.80 & 5.08 & 17.82 & 51.95\\
                IN1K & ViT-B & 87M & 224x224 & 196 & 26.66 & 15.41 & 19.12 & 4.14 & 13.92 & 51.36\\
                \bottomrule
            \end{tabular}
            }
    \end{small}
  \end{center}
  \vskip -0.1in            
\end{table*}

\begin{table*}[t] 
    \caption{MaxViT VQA benchmarks}
    \vskip -0.1in
    \label{tab:exhaustive-MaxViT-vqa}
      \begin{center}
        \begin{small}
            \resizebox{\textwidth}{!}{%
            \begin{tabular}{c c c c c | c c c c c c c | c} 
                \toprule
                Pretrained & Visual & Encoder & Image & Vision & \multirow{2}{*}{VQA-v2} & \multirow{2}{*}{GQA} & \multirow{2}{*}{VizWiz} & TextVQA & \multirow{2}{*}{TextVQA} & \multirow{2}{*}{POPE} & \multirow{2}{*}{TallyQA} & Weighted \\
                Dataset & Encoder & Size & Size & Token \# &  &  &  & +OCR &  &  &  & VQA \\
                \midrule
                IN21K & MaxViT-B & 119M & 224x224 & 196 & 62.67 & 49.98 & 45.92 & 44.08 & 12.38 & 76.46 & 53.58 & 59.88\\
                IN21K & MaxViT-L & 212M & 224x224 & 196 & 65.50 & 52.03 & 47.31 & 44.30 & 12.49 & 80.59 & 53.62 & 62.23\\
                IN21K & MaxViT-XL & 475M & 224x224 & 196 & 65.55 & 51.61 & 50.14 & 44.13 & 12.58 & 80.07 & \underline{54.78} & 62.43\\
                IN21K $\to$ IN1K & MaxViT-B & 119M & 384x384 & 576 & 63.44 & 50.92 & 50.41 & 44.38 & 12.71 & 76.72 & 54.53 & 60.71\\
                IN21K $\to$ IN1K & MaxViT-L & 212M & 384x384 & 576 & 67.15 & 52.82 & \textbf{53.89} & 43.72 & \textbf{13.28} & 82.30 & \textbf{55.45} & \textbf{63.89}\\
                IN21K $\to$ IN1K & MaxViT-XL & 475M & 384x384 & 576 & 66.37 & 52.53 & 49.11 & 44.16 & 12.89 & 81.46 & 53.58 & 62.95\\
                IN21K $\to$ IN1K & MaxViT-B & 119M & 512x512 & 1024 & 63.42 & 51.07 & 49.87 & 44.10 & 12.53 & 77.67 & 54.35 & 60.69\\
                IN21K $\to$ IN1K & MaxViT-L & 212M & 512x512 & 1024 & \underline{67.27} & 52.70 & \underline{51.91} & \textbf{44.71} & \underline{13.09} & \underline{83.11} & 54.38 & \underline{63.84}\\
                IN21K $\to$ IN1K & MaxViT-XL & 475M & 512x512 & 1024 & \textbf{67.34} & 52.72 & 51.08 & 44.16 & 12.95 & \textbf{83.15} & 52.83 & 63.67\\
                \midrule
                \arrayrulecolor{black}
                IN1K & MaxViT-T & 31M & 384x384 & 576 & 64.04 & \underline{53.04} & 48.00 & \underline{44.69} & 11.71 & 79.86 & 53.30 & 61.13\\
                IN1K & MaxViT-S & 69M & 384x384 & 576 & 61.31 & 49.85 & 46.73 & 44.19 & 11.79 & 77.57 & 52.05 & 58.70\\
                IN1K & MaxViT-B & 119M & 384x384 & 576 & 63.90 & 52.76 & 46.61 & 43.95 & 12.27 & 81.23 & 52.83 & 60.97\\
                IN1K & MaxViT-L & 212M & 384x384 & 576 & 62.90 & 51.26 & 49.94 & 43.80 & 12.41 & 81.00 & 49.25 & 59.73\\
                IN1K & MaxViT-T & 31M & 512x512 & 1024 & 64.25 & 52.84 & 48.14 & 44.40 & 11.84 & 79.24 & 51.93 & 61.08\\
                IN1K & MaxViT-S & 69M & 512x512 & 1024 & 63.51 & 51.26 & 46.29 & 44.25 & 11.92 & 79.36 & 54.01 & 60.71\\
                IN1K & MaxViT-B & 119M & 512x512 & 1024 & 64.53 & \textbf{53.14} & 47.92 & 44.10 & 12.09 & 82.90 & 53.06 & 61.56\\
                IN1K & MaxViT-L & 212M & 512x512 & 1024 & 63.19 & 51.17 & 46.09 & 43.63 & 12.41 & 81.30 & 48.87 & 59.84\\
                \midrule
                \arrayrulecolor{black}
                IN1K & MaxViT-T & 31M & 224x224 & 196 & 61.08 & 49.43 & 47.58 & 44.31 & 12.02 & 76.77 & 53.85 & 58.75\\
                IN1K & MaxViT-S & 69M & 224x224 & 196 & 59.90 & 48.94 & 41.50 & 44.51 & 11.82 & 76.29 & 51.42 & 57.42\\
                IN1K & MaxViT-B & 119M & 224x224 & 196 & 60.10 & 49.72 & 45.00 & 44.59 & 12.29 & 76.26 & 50.94 & 57.60\\
                IN1K & MaxViT-L & 212M & 224x224 & 196 & 60.07 & 49.55 & 46.42 & 43.67 & 12.23 & 76.46 & 48.86 & 57.30\\
                \bottomrule
            \end{tabular}
            }
    \end{small}
  \end{center}
  \vskip -0.1in            
\end{table*}

\begin{table*}[t] 
    \caption{MaxViT localization benchmarks}
    \vskip -0.1in
    \label{tab:exhaustive-MaxViT-loc}
      \begin{center}
        \begin{small}
            \resizebox{\textwidth}{!}{%
            \begin{tabular}{c c c c c | c c c c | c || c} 
                \toprule
                Pretrained & Visual & Encoder & Image & Vision & \multirow{2}{*}{RefCOCO} & \multirow{2}{*}{RefCOCO+} & \multirow{2}{*}{RefCOCOg} & \multirow{2}{*}{OCID-Ref} & Weighted & Weighted \\
                Dataset & Encoder & Size & Size & Token \# &  &  &  &  & Loc. & Overall \\
                \midrule
                IN21K & MaxViT-B & 119M & 224x224 & 196 & 23.04 & 12.72 & 16.44 & 2.56 & 11.46 & 53.38\\
                IN21K & MaxViT-L & 212M & 224x224 & 196 & 26.27 & 15.83 & 19.53 & 4.94 & 14.30 & 55.79\\
                IN21K & MaxViT-XL & 475M & 224x224 & 196 & 24.43 & 13.70 & 18.34 & 4.46 & 13.02 & 55.79\\
                IN21K $\to$ IN1K & MaxViT-B & 119M & 384x384 & 576 & 21.22 & 12.01 & 14.56 & 2.87 & 10.77 & 54.00\\
                IN21K $\to$ IN1K & MaxViT-L & 212M & 384x384 & 576 & 26.46 & 15.92 & 19.36 & 5.79 & 14.70 & 57.28\\
                IN21K $\to$ IN1K & MaxViT-XL & 475M & 384x384 & 576 & 24.67 & 14.46 & 18.32 & 3.67 & 12.93 & 56.23\\
                IN21K $\to$ IN1K & MaxViT-B & 119M & 512x512 & 1024 & 21.72 & 12.49 & 15.26 & 2.58 & 10.97 & 54.01\\
                IN21K $\to$ IN1K & MaxViT-L & 212M & 512x512 & 1024 & 26.45 & 15.50 & 20.26 & 5.22 & 14.46 & 57.21\\
                IN21K $\to$ IN1K & MaxViT-XL & 475M & 512x512 & 1024 & 26.67 & 15.66 & 20.20 & 5.33 & 14.59 & 57.07\\
                \midrule
                \arrayrulecolor{black}
                IN1K & MaxViT-T & 31M & 384x384 & 576 & \textbf{55.01} & \underline{43.06} & \underline{47.75} & \underline{22.30} & \underline{37.97} & \underline{58.02}\\
                IN1K & MaxViT-S & 69M & 384x384 & 576 & 25.68 & 14.97 & 18.22 & 4.08 & 13.46 & 52.62\\
                IN1K & MaxViT-B & 119M & 384x384 & 576 & 28.18 & 18.45 & 20.61 & 6.08 & 15.98 & 54.92\\
                IN1K & MaxViT-L & 212M & 384x384 & 576 & 23.92 & 13.65 & 17.12 & 5.36 & 13.12 & 53.46\\
                IN1K & MaxViT-T & 31M & 512x512 & 1024 & \underline{54.54} & \textbf{43.28} & \textbf{48.24} & \textbf{23.42} & \textbf{38.42} & \textbf{58.03}\\
                IN1K & MaxViT-S & 69M & 512x512 & 1024 & 35.07 & 23.51 & 28.10 & 9.29 & 20.99 & 55.37\\
                IN1K & MaxViT-B & 119M & 512x512 & 1024 & 33.23 & 22.73 & 25.14 & 8.34 & 19.64 & 55.92\\
                IN1K & MaxViT-L & 212M & 512x512 & 1024 & 25.60 & 15.34 & 18.83 & 5.90 & 14.34 & 53.72\\
                \midrule
                \arrayrulecolor{black}
                IN1K & MaxViT-T & 31M & 224x224 & 196 & 29.44 & 17.29 & 22.10 & 5.17 & 15.79 & 52.98\\
                IN1K & MaxViT-S & 69M & 224x224 & 196 & 25.57 & 14.41 & 17.28 & 4.38 & 13.32 & 51.49\\
                IN1K & MaxViT-B & 119M & 224x224 & 196 & 22.02 & 12.67 & 15.28 & 3.36 & 11.41 & 51.39\\
                IN1K & MaxViT-L & 212M & 224x224 & 196 & 21.15 & 12.01 & 14.77 & 2.94 & 10.81 & 51.05\\
                \bottomrule
            \end{tabular}
            }
    \end{small}
  \end{center}
  \vskip -0.1in            
\end{table*}

\begin{table*}[t] 
    \caption{MambaVision VQA benchmarks}
    \vskip -0.1in
    \label{tab:exhaustive-MambaVision-vqa}
      \begin{center}
        \begin{small}
            \resizebox{\textwidth}{!}{%
            \begin{tabular}{c c c c c | c c c c c c c | c} 
                \toprule
                Pretrained & Visual & Encoder & Image & Vision & \multirow{2}{*}{VQA-v2} & \multirow{2}{*}{GQA} & \multirow{2}{*}{VizWiz} & TextVQA & \multirow{2}{*}{TextVQA} & \multirow{2}{*}{POPE} & \multirow{2}{*}{TallyQA} & Weighted \\
                Dataset & Encoder & Size & Size & Token \# &  &  &  & +OCR &  &  &  & VQA \\
                \midrule
                IN21K & MambaVision-B & 98M & 224x224 & 196 & 49.42 & 42.07 & 41.98 & 42.71 & 10.36 & 58.90 & 42.05 & 47.50\\
                IN21K & MambaVision-L & 228M & 224x224 & 196 & 52.06 & 43.27 & 39.10 & 43.12 & 10.64 & 65.22 & 43.49 & 49.87\\
                IN21K & MambaVision-L2 & 242M & 512x512 & 1024 & 50.80 & 41.95 & 41.57 & 42.79 & 10.50 & 62.97 & 41.93 & 48.63\\
                IN21K & MambaVision-L3 & 740M & 256x256 & 256 & 52.18 & 43.58 & 36.27 & 42.94 & 10.66 & 65.13 & 38.51 & 49.26\\
                IN21K & MambaVision-L3 & 740M & 512x512 & 1024 & 51.50 & 42.62 & 40.91 & 43.40 & 10.45 & 61.98 & 43.19 & 49.32\\
                \midrule
                \arrayrulecolor{black}
                IN1K & MambaVision-T & 32M & 224x224 & 196 & 56.85 & 47.43 & \underline{46.16} & \underline{43.74} & 11.14 & 71.34 & 46.89 & 54.38\\
                IN1K & MambaVision-T2 & 35M & 224x224 & 196 & 56.69 & 47.14 & 43.17 & 43.31 & 10.82 & 69.58 & 43.74 & 53.71\\
                IN1K & MambaVision-S & 50M & 224x224 & 196 & 58.10 & 48.05 & 40.91 & 43.08 & 11.36 & 72.03 & 49.48 & 55.61\\
                IN1K & MambaVision-B & 98M & 224x224 & 196 & 59.14 & \underline{49.23} & 44.21 & 43.33 & \textbf{11.43} & 73.05 & 49.53 & 56.53\\
                IN1K & MambaVision-L & 228M & 224x224 & 196 & \underline{59.26} & \textbf{49.56} & 42.13 & \textbf{43.75} & \underline{11.41} & \textbf{74.46} & \textbf{51.70} & \textbf{56.94}\\
                IN1K & MambaVision-L2 & 242M & 224x224 & 196 & \textbf{59.27} & 48.83 & \textbf{48.91} & 43.30 & 11.16 & \underline{74.15} & \underline{50.70} & \underline{56.86}\\
                \bottomrule
            \end{tabular}
            }
    \end{small}
  \end{center}
  \vskip -0.1in            
\end{table*}

\begin{table*}[t] 
    \caption{MambaVision localization benchmarks }
    \vskip -0.1in
    \label{tab:exhaustive-MambaVision-loc}
      \begin{center}
        \begin{small}
            \resizebox{\textwidth}{!}{%
            \begin{tabular}{c c c c c | c c c c | c || c} 
                \toprule
                Pretrained & Visual & Encoder & Image & Vision & \multirow{2}{*}{RefCOCO} & \multirow{2}{*}{RefCOCO+} & \multirow{2}{*}{RefCOCOg} & \multirow{2}{*}{OCID-Ref} & Weighted & Weighted \\
                Dataset & Encoder & Size & Size & Token \# &  &  &  &  & Loc. & Overall \\
                \midrule
                IN21K & MambaVision-B & 98M & 224x224 & 196 & 19.44 & 10.68 & 12.99 & 3.52 & 10.12 & 42.48\\
                IN21K & MambaVision-L & 228M & 224x224 & 196 & 22.52 & 11.91 & 14.69 & 3.91 & 11.50 & 44.72\\
                IN21K & MambaVision-L2 & 242M & 512x512 & 1024 & 21.80 & 12.46 & 14.17 & 3.45 & 11.22 & 43.60\\
                IN21K & MambaVision-L3 & 740M & 256x256 & 256 & 21.39 & 11.89 & 13.93 & 3.71 & 11.06 & 44.13\\
                IN21K & MambaVision-L3 & 740M & 512x512 & 1024 & 21.27 & 11.61 & 13.46 & 3.66 & 10.89 & 44.15\\
                \midrule
                \arrayrulecolor{black}
                IN1K & MambaVision-T & 32M & 224x224 & 196 & 34.59 & 24.01 & 27.14 & 9.52 & 20.98 & 49.89\\
                IN1K & MambaVision-T2 & 35M & 224x224 & 196 & 35.78 & 24.69 & 27.61 & 9.72 & 21.56 & 49.39\\
                IN1K & MambaVision-S & 50M & 224x224 & 196 & 44.65 & 33.06 & 37.48 & 13.21 & 28.22 & 51.93\\
                IN1K & MambaVision-B & 98M & 224x224 & 196 & \textbf{48.52} & \underline{35.57} & \textbf{40.58} & 15.84 & 31.17 & 53.12\\
                IN1K & MambaVision-L & 228M & 224x224 & 196 & \underline{48.19} & \textbf{35.82} & 40.26 & \textbf{16.79} & \textbf{31.51} & \textbf{53.52}\\
                IN1K & MambaVision-L2 & 242M & 224x224 & 196 & 47.91 & 35.41 & \underline{40.28} & \underline{16.48} & \underline{31.22} & \underline{53.42}\\
                \bottomrule
            \end{tabular}
            }
    \end{small}
  \end{center}
  \vskip -0.1in            
\end{table*}

\begin{table*}[t] 
    \caption{Vim VQA benchmarks}
    \vskip -0.1in
    \label{tab:exhaustive-Vim-vqa}
      \begin{center}
        \begin{small}
            \resizebox{\textwidth}{!}{%
            \begin{tabular}{c c c c c | c c c c c c c | c} 
                \toprule
                Pretrained & Visual & Encoder & Image & Vision & \multirow{2}{*}{VQA-v2} & \multirow{2}{*}{GQA} & \multirow{2}{*}{VizWiz} & TextVQA & \multirow{2}{*}{TextVQA} & \multirow{2}{*}{POPE} & \multirow{2}{*}{TallyQA} & Weighted \\
                Dataset & Encoder & Size & Size & Token \# &  &  &  & +OCR &  &  &  & VQA \\
                \midrule
                IN1K & Vim-T & 7M & 224x224 & 196 & 55.83 & 46.93 & 39.39 & \underline{43.82} & 11.67 & 72.91 & 45.69 & 53.40\\
                IN1K & Vim-T-ft & 7M & 224x224 & 729 & 56.82 & 47.62 & 41.51 & 43.56 & 11.76 & \underline{74.69} & 47.73 & 54.52\\
                IN1K & Vim-S & 26M & 224x224 & 196 & 56.84 & 47.73 & 40.47 & 43.81 & 11.58 & 73.58 & \underline{49.58} & 54.74\\
                IN1K & Vim-S-ft & 26M & 224x224 & 729 & \underline{57.68} & \underline{48.55} & \textbf{45.68} & 43.78 & \underline{11.89} & 74.31 & 45.96 & \underline{55.02}\\
                IN1K & Vim-B & 98M & 224x224 & 196 & \textbf{62.93} & \textbf{52.11} & \underline{45.52} & \textbf{44.81} & \textbf{12.50} & \textbf{78.97} & \textbf{55.07} & \textbf{60.46}\\
                \bottomrule
            \end{tabular}
            }
    \end{small}
  \end{center}
  \vskip -0.1in            
\end{table*}

\begin{table*}[t] 
    \caption{Vim localization benchmarks}
    \vskip -0.1in
    \label{tab:exhaustive-Vim-loc}
      \begin{center}
        \begin{small}
            \resizebox{\textwidth}{!}{%
            \begin{tabular}{c c c c c | c c c c | c || c} 
                \toprule
                Pretrained & Visual & Encoder & Image & Vision & \multirow{2}{*}{RefCOCO} & \multirow{2}{*}{RefCOCO+} & \multirow{2}{*}{RefCOCOg} & \multirow{2}{*}{OCID-Ref} & Weighted & Weighted \\
                Dataset & Encoder & Size & Size & Token \# &  &  &  &  & Loc. & Overall \\
                \midrule
                IN1K & Vim-T & 7M & 224x224 & 196 & 21.35 & 12.42 & 14.99 & 3.51 & 11.21 & 47.73\\
                IN1K & Vim-T-ft & 7M & 224x224 & 729 & 21.28 & 12.30 & 15.13 & \underline{3.96} & 11.37 & 48.72\\
                IN1K & Vim-S & 26M & 224x224 & 196 & 22.33 & 12.60 & \underline{15.32} & 2.70 & 11.20 & 48.88\\
                IN1K & Vim-S-ft & 26M & 224x224 & 729 & \underline{22.99} & \underline{13.09} & 15.01 & 3.57 & \underline{11.80} & \underline{49.21}\\
                IN1K & Vim-B & 98M & 224x224 & 196 & \textbf{44.62} & \textbf{34.10} & \textbf{38.54} & \textbf{13.17} & \textbf{28.56} & \textbf{56.17}\\
                \bottomrule
            \end{tabular}
            }
    \end{small}
  \end{center}
  \vskip -0.1in            
\end{table*}

\begin{table*}[t] 
    \caption{ViTDet VQA benchmarks}
    \vskip -0.1in
    \label{tab:exhaustive-ViTDet-vqa}
      \begin{center}
        \begin{small}
            \resizebox{\textwidth}{!}{%
            \begin{tabular}{c c c c c | c c c c c c c | c} 
                \toprule
                Pretrained & Visual & Encoder & Image & Vision & \multirow{2}{*}{VQA-v2} & \multirow{2}{*}{GQA} & \multirow{2}{*}{VizWiz} & TextVQA & \multirow{2}{*}{TextVQA} & \multirow{2}{*}{POPE} & \multirow{2}{*}{TallyQA} & Weighted \\
                Dataset & Encoder & Size & Size & Token \# &  &  &  & +OCR &  &  &  & VQA \\
                \midrule
                IN1K $\to$ COCO & ViTDet-B & 111M & 1024x1024 & 4096 & 65.83 & \underline{53.61} & \textbf{50.58} & 43.75 & 11.60 & 84.46 & 55.96 & 63.00\\
                IN1K $\to$ COCO & ViTDet-L & 331M & 1024x1024 & 4096 & 60.00 & 48.59 & 44.78 & 43.88 & \underline{11.78} & 78.14 & 51.94 & 57.64\\
                IN1K $\to$ COCO & ViTDet-H & 662M & 1024x1024 & 4096 & 64.43 & 52.62 & \underline{48.99} & \underline{43.93} & 11.50 & 84.29 & 55.95 & 61.89\\
                \midrule
                \arrayrulecolor{black}
                IN1K $\to$ COCO & ViTDet-B(f) & 111M & 1024x1024 & 4096 & 63.68 & 52.14 & 46.14 & 43.64 & 11.69 & 82.58 & 53.52 & 60.89\\
                IN1K $\to$ COCO & ViTDet-L(f) & 331M & 1024x1024 & 4096 & \underline{66.43} & 53.54 & 45.52 & 43.76 & 11.58 & \underline{84.82} & \underline{57.24} & \underline{63.55}\\
                IN1K $\to$ COCO & ViTDet-H(f) & 662M & 1024x1024 & 4096 & \textbf{66.89} & \textbf{54.22} & 41.01 & \textbf{44.04} & \textbf{11.82} & \textbf{87.03} & \textbf{58.08} & \textbf{64.05}\\
                \bottomrule
            \end{tabular}
            }
    \end{small}
  \end{center}
  \vskip -0.1in            
\end{table*}

\begin{table*}[t] 
    \caption{ViTDet localization benchmarks}
    \vskip -0.1in
    \label{tab:exhaustive-ViTDet-loc}
      \begin{center}
        \begin{small}
            \resizebox{\textwidth}{!}{%
            \begin{tabular}{c c c c c | c c c c | c || c} 
                \toprule
                Pretrained & Visual & Encoder & Image & Vision & \multirow{2}{*}{RefCOCO} & \multirow{2}{*}{RefCOCO+} & \multirow{2}{*}{RefCOCOg} & \multirow{2}{*}{OCID-Ref} & Weighted & Weighted \\
                Dataset & Encoder & Size & Size & Token \# &  &  &  &  & Loc. & Overall \\
                \midrule
                IN1K $\to$ COCO & ViTDet-B & 111M & 1024x1024 & 4096 & \underline{66.03} & 51.17 & 58.17 & 22.37 & 43.74 & 60.42\\
                IN1K $\to$ COCO & ViTDet-L & 331M & 1024x1024 & 4096 & 24.62 & 13.79 & 17.44 & 4.62 & 13.05 & 51.65\\
                IN1K $\to$ COCO & ViTDet-H & 662M & 1024x1024 & 4096 & 56.41 & 40.48 & 50.25 & 16.01 & 35.38 & 58.33\\
                \midrule
                \arrayrulecolor{black}
                IN1K $\to$ COCO & ViTDet-B(f) & 111M & 1024x1024 & 4096 & 46.03 & 33.49 & 39.40 & 11.72 & 28.26 & 56.51\\
                IN1K $\to$ COCO & ViTDet-L(f) & 331M & 1024x1024 & 4096 & 65.73 & \underline{52.09} & \underline{58.86} & \underline{23.87} & \underline{44.58} & \underline{61.00}\\
                IN1K $\to$ COCO & ViTDet-H(f) & 662M & 1024x1024 & 4096 & \textbf{69.37} & \textbf{54.27} & \textbf{61.97} & \textbf{26.51} & \textbf{47.40} & \textbf{61.81}\\
                \bottomrule
            \end{tabular}
            }
    \end{small}
  \end{center}
  \vskip -0.1in            
\end{table*}

\begin{table*}[t] 
    \caption{ViT-Adapter VQA benchmarks}
    \vskip -0.1in
    \label{tab:exhaustive-ViT-Adapter-vqa}
      \begin{center}
        \begin{small}
            \resizebox{\textwidth}{!}{%
            \begin{tabular}{c c c c c | c c c c c c c | c} 
                \toprule
                Pretrained & Visual & Encoder & Image & Vision & \multirow{2}{*}{VQA-v2} & \multirow{2}{*}{GQA} & \multirow{2}{*}{VizWiz} & TextVQA & \multirow{2}{*}{TextVQA} & \multirow{2}{*}{POPE} & \multirow{2}{*}{TallyQA} & Weighted \\
                Dataset & Encoder & Size & Size & Token \# &  &  &  & +OCR &  &  &  & VQA \\
                \midrule
                Multi-Modal $\to$ ADE20K & Perceiver & 364M & 512x512 & 256 & 65.62 & 54.63 & 46.25 & 44.26 & 12.25 & 83.79 & 56.62 & 62.92\\
                IN22K $\to$ ADE20K & BeiT-L & 451M & 640x640 & 400 & 68.78 & 56.52 & 48.06 & \textbf{44.75} & 13.01 & 86.88 & 58.24 & 65.70\\
                IN22K $\to$ ADE20K & BeiT-L & 568M & 640x640 & 400 & \underline{69.57} & 57.12 & \underline{51.95} & 44.03 & \underline{13.14} & 86.57 & 58.64 & 66.41\\
                IN22K+COCO $\to$ ADE20K & BeiT-L & 571M & 896x896 & 784 & \underline{69.57} & \underline{57.47} & 49.73 & 44.58 & 12.89 & \underline{88.09} & \underline{59.03} & \underline{66.49}\\
                IN22K+COCO $\to$ ADE20K & BeiTv2-L & 571M & 896x896 & 784 & \textbf{70.84} & \textbf{58.58} & 49.84 & 44.28 & \textbf{13.39} & \textbf{88.24} & \textbf{61.32} & \textbf{67.80}\\
                \midrule
                \arrayrulecolor{black}
                IN22K $\to$ ADE20K & AugReg-T & 36M & 512x512 & 256 & 58.52 & 50.25 & 47.98 & 43.92 & 11.53 & 76.08 & 52.32 & 56.65\\
                IN22K $\to$ ADE20K & AugReg-B & 134M & 512x512 & 256 & 61.51 & 52.68 & 46.41 & \underline{44.65} & 11.99 & 78.28 & 54.20 & 59.29\\
                IN22K $\to$ ADE20K & AugReg-L & 364M & 512x512 & 256 & 65.92 & 55.60 & 50.37 & 44.42 & 12.32 & 83.23 & 54.98 & 63.01\\
                \midrule
                \arrayrulecolor{black}
                IN1K $\to$ ADE20K & DeiT-S & 58M & 512x512 & 256 & 61.56 & 52.37 & \textbf{52.16} & 44.25 & 11.74 & 79.82 & 53.60 & 59.36\\
                IN1K $\to$ ADE20K & DeiT-B & 134M & 512x512 & 256 & 63.37 & 53.32 & 49.40 & 43.98 & 11.97 & 81.64 & 53.11 & 60.69\\
                \bottomrule
            \end{tabular}
            }
    \end{small}
  \end{center}
  \vskip -0.1in            
\end{table*}

\begin{table*}[t] 
    \caption{ViT-Adapter localization benchmarks}
    \vskip -0.1in
    \label{tab:exhaustive-ViT-Adapter-loc}
      \begin{center}
        \begin{small}
            \resizebox{\textwidth}{!}{%
            \begin{tabular}{c c c c c | c c c c | c || c} 
                \toprule
                Pretrained & Visual & Encoder & Image & Vision & \multirow{2}{*}{RefCOCO} & \multirow{2}{*}{RefCOCO+} & \multirow{2}{*}{RefCOCOg} & \multirow{2}{*}{OCID-Ref} & Weighted & Weighted \\
                Dataset & Encoder & Size & Size & Token \# &  &  &  &  & Loc. & Overall \\
                \midrule
                Multi-Modal $\to$ ADE20K & Perceiver & 364M & 512x512 & 256 & \textbf{63.31} & \textbf{51.55} & \textbf{55.88} & \underline{20.81} & \underline{42.29} & 60.14\\
                IN22K $\to$ ADE20K & BeiT-L & 451M & 640x640 & 400 & 52.99 & 42.75 & 47.81 & 16.94 & 35.22 & 61.61\\
                IN22K $\to$ ADE20K & BeiT-L & 568M & 640x640 & 400 & 56.09 & 46.49 & 51.45 & 18.60 & 37.94 & 62.58\\
                IN22K+COCO $\to$ ADE20K & BeiT-L & 571M & 896x896 & 784 & 55.53 & 46.07 & 49.43 & 18.51 & 37.45 & \underline{62.59}\\
                IN22K+COCO $\to$ ADE20K & BeiTv2-L & 571M & 896x896 & 784 & \underline{60.34} & \underline{49.29} & \underline{53.33} & \textbf{24.69} & \textbf{42.34} & \textbf{64.38}\\
                \midrule
                \arrayrulecolor{black}
                IN22K $\to$ ADE20K & AugReg-T & 36M & 512x512 & 256 & 41.19 & 29.14 & 34.17 & 11.32 & 25.31 & 52.44\\
                IN22K $\to$ ADE20K & AugReg-B & 134M & 512x512 & 256 & 48.32 & 36.36 & 41.44 & 15.54 & 31.29 & 55.53\\
                IN22K $\to$ ADE20K & AugReg-L & 364M & 512x512 & 256 & 56.58 & 45.16 & 50.47 & 20.28 & 38.32 & 59.69\\
                \midrule
                \arrayrulecolor{black}
                IN1K $\to$ ADE20K & DeiT-S & 58M & 512x512 & 256 & 49.52 & 37.07 & 42.46 & 15.35 & 31.78 & 55.65\\
                IN1K $\to$ ADE20K & DeiT-B & 134M & 512x512 & 256 & 52.17 & 39.78 & 45.61 & 16.22 & 33.77 & 57.07\\
                \bottomrule
            \end{tabular}
            }
    \end{small}
  \end{center}
  \vskip -0.1in            
\end{table*}

\end{document}